\PassOptionsToPackage{table}{xcolor}
\documentclass[sigconf]{acmart}

\AtBeginDocument{%
  }

\usepackage{hyperref}       
\usepackage{url}            
\usepackage{booktabs}       
\usepackage{amsfonts}       
\usepackage{nicefrac}       
\usepackage{microtype}      
\usepackage[table]{xcolor} 
\usepackage{makecell}
\usepackage{enumitem}
\usepackage{array}
\usepackage{wrapfig}
\usepackage{rotating}
\usepackage{amsmath}
\usepackage{subfigure}
\usepackage{tcolorbox}
\usepackage{caption}
\usepackage{tabularx}
\usepackage{multirow}
\usepackage{xcolor}
\usepackage{twemojis}
\usepackage{titlesec}
\titleformat{\section}{\normalfont\Large\bfseries\MakeUppercase} 
  {\thesection}{1em}{}
\newcommand{\mynl}{\textcolor{gray}{\textbackslash n}}
\definecolor{myshadow}{rgb}{0.901,0.901,0.901}

\copyrightyear{2025}
\acmYear{2025}
\setcopyright{acmlicensed}
\acmConference[KDD '25] {Proceedings of the 31st ACM SIGKDD Conference on Knowledge Discovery and Data Mining V.1}{August 3--7, 2025}{Toronto, ON, Canada.}
\acmBooktitle{Proceedings of the 31st ACM SIGKDD Conference on Knowledge Discovery and Data Mining V.1 (KDD '25), August 3--7, 2025, Toronto, ON, Canada}
\acmISBN{979-8-4007-1245-6/25/08}
\acmDOI{	https://doi.org/10.1145/3690624.3709256}
\begin{document}
\title{Can Large Language Models Improve the Adversarial Robustness of Graph Neural Networks?}

\author{Zhongjian Zhang$^{1\ast}$, Xiao Wang$^{2\ast}$, Huichi Zhou$^3$, Yue Yu$^1$\\Mengmei Zhang$^4$, Cheng Yang$^1$ and Chuan Shi$^{1\dagger}$}
\thanks{$\ast$ Both authors contributed equally to this research.\\$\dagger$ Corresponding author.}
\affiliation{$^1$Beijing University of Posts and Telecommunications, $^2$Beihang University\\$^3$Imperial College London, $^4$China Telecom Bestpay\\\twemoji{e-mail} {zhangzj@bupt.edu.cn, xiao\_wang@buaa.edu.cn, h.zhou24@imperial.ac.uk, yuyue1218@bupt.edu.cn\\zhangmengmei@bestpay.com.cn, yangcheng@bupt.edu.cn, shichuan@bupt.edu.cn}  \country{}
}
\renewcommand{\shortauthors}{Zhongjian Zhang, Xiao Wang and Huichi Zhou et al.}

\begin{abstract}
Graph neural networks (GNNs) are vulnerable to adversarial attacks, especially for topology perturbations, and many methods that improve the robustness of GNNs have received considerable attention. Recently, we have witnessed the significant success of large language models (LLMs), leading many to explore the great potential of LLMs on GNNs. However, they mainly focus on improving the performance of GNNs by utilizing LLMs to enhance the node features. Therefore, we ask: \textit{Will the robustness of GNNs also be enhanced with the powerful understanding and inference capabilities of LLMs?} By presenting the empirical results, we find that despite that LLMs can improve the robustness of GNNs, there is still an average decrease of 23.1\% in accuracy, implying that the GNNs remain extremely vulnerable against topology attacks. Therefore, another question is \textit{how to extend the capabilities of LLMs on graph adversarial robustness}. In this paper, we propose an LLM-based robust graph structure inference framework, LLM4RGNN, which distills the inference capabilities of GPT-4 into a local LLM for identifying malicious edges and an LM-based edge predictor for finding missing important edges, so as to recover a robust graph structure. Extensive experiments demonstrate that LLM4RGNN consistently improves the robustness across various GNNs. Even in some cases where the perturbation ratio increases to 40\%, the accuracy of GNNs is still better than that on the clean graph. The source code can be found in \href{https://github.com/zhongjian-zhang/LLM4RGNN}{https://github.com/zhongjian-zhang/LLM4RGNN}.
\end{abstract}

\ccsdesc[300]{Mathematics of computing~Graph algorithms}
\ccsdesc[300]{Security and privacy~Social network security and privacy}

\keywords{Graph Neural Networks; Large Language Models; Adversarial Robustness; Structure Learning}

\maketitle
\section{Introduction}
Graph neural networks (GNNs), as representative graph machine learning methods, effectively utilize their message-passing mechanism to extract useful information and learn high-quality representations from graph data~\cite{kipf2016semi, velivckovic2017graph,wang2017community,zhang2024endowing,du2021multi}.
Despite great success, a host of studies have shown that GNNs are vulnerable to adversarial attacks~\cite{sun2022adversarial, li2022revisiting, mujkanovic2022defenses, waniek2018hiding, jin2021adversarial}, especially for topology attacks~\cite{zugner_adversarial_2019, zugner2018adversarial, xu2019topology}, where slightly perturbing the graph structure can lead to a dramatic decrease in the performance. Such vulnerability poses significant challenges for applying GNNs to real-world applications, especially in security-critical scenarios such as finance networks~\cite{wang2021review} or medical networks~\cite{mao2019medgcn}. 

Threatened by adversarial attacks, several attempts have been made to build robust GNNs, which can be mainly divided into model-centric and data-centric defenses~\cite{zheng2021grb, guo2024data}.
From the model-centric perspective, defenders can improve robustness through model enhancement, either by robust training schemes~\cite{li2023boosting, gosch2024adversarial} or new model architectures~\cite{zhu2019robust, jin2021node, zhao2024adversarial}. 
In contrast, data-centric defenses typically focus on flexible data processing to improve the robustness of GNNs. Treating the attacked topology as noisy, defenders primarily purify graph structures 
by calculating various similarities between node embeddings~\cite{wu2019adversarial, jin2020graph, zhang2020gnnguard, li2022reliable}. 
The above methods have received considerable attention in enhancing the robustness of GNNs.

Recently, large language models (LLMs), such as GPT-4~\cite{achiam2023gpt}, have demonstrated expressive capabilities in understanding and inferring complex texts, revolutionizing the fields of natural language processing~\cite{zhao2023survey, zhou2024evaluating}, computer vision~\cite{yin2023survey} and graph~\cite{liu2023towards}. The performance of GNNs can be greatly improved by utilizing LLMs to enhance the node features~\cite{he2023harnessing, liu2023one, chen2024exploring}. However, one question remains largely unknown: \textit{Considering the powerful understanding and inference capabilities of LLMs, will LLMs enhance or weaken the adversarial robustness of GNNs to a certain extent?}
Answering this question not only helps explore the potential capabilities of LLMs on graphs, but also provides a new perspective for the adversarial robustness problem on graphs.

Here, we empirically investigate the robustness of GNNs combining six LLMs/LMs (language models), namely OFA-Llama2-7B~\cite{liu2023one}, OFA-SBert~\cite{liu2023one}, TAPE~\cite{he2023harnessing}, GCN-Llama-7B~\cite{chen2024exploring}, GCN-e5-large~\cite{chen2024exploring}, and GCN-SBert~\cite{chen2024exploring}, against Mettack~\cite{zugner_adversarial_2019} with a 20\% perturbation rate on Cora~\cite{mccallum2000automating} and PubMed~\cite{sen2008collective} datasets. As shown in Figure~\ref{fig: pre}, the results clearly show that these models suffer from a maximum accuracy decrease of 37.9\% and an average of 23.1\%, while vanilla GCN~\cite{kipf2016semi} experiences a maximum accuracy decrease of 39.1\% and an average of 35.5\%. 
It demonstrates that these models remain extremely vulnerable to topology perturbations (more details refer to Section~\ref{sec: preliminary}). Consequently, another question naturally arises: \textit{How to extend the capabilities of LLMs to improve graph adversarial robustness?} 
This problem is non-trivial because graph adversarial attacks typically perturb the graph structures, while the capabilities of LLMs usually focus on text processing. Considering that graph structures involve complex interactions among a large number of nodes, how to efficiently explore the inference capabilities of LLMs on perturbed structures is a significant challenge.

In this paper, we propose an LLM-based robust graph structure inference framework, called LLM4RGNN, which efficiently utilizes LLMs to purify the perturbed structure, improving the adversarial robustness. Specifically, based on an open-source and clean graph structure, we design a prompt template that enables GPT-4~\cite{achiam2023gpt} to infer how malicious an edge is and provide analysis, to construct an instruction dataset. This dataset is used to fine-tune a local LLM (e.g., Mistral-7B~\cite{jiang2023mistral} or Llama3-8B~\cite{touvron2023llama}), so that the inference capability of GPT-4 can be distilled into the local LLM.
When given a new attacked graph structure, we first utilize the local LLM to identify malicious edges. By treating identification results as edge labels, we further distill the inference capability from the local LLM to an LM-based edge predictor, to find missing important edges. Finally, purifying the graph structure by removing malicious edges and adding important edges, makes GNNs more robust. 
Our contributions can be summarized four-fold:
\begin{itemize}[leftmargin=*]
\item To the best of our knowledge, we are the first to explore the potential of LLMs on the graph adversarial robustness. Moreover, we verify the vulnerability of GNNs even with the powerful understanding and inference capabilities of LLMs.
\item We propose a novel LLM-based robust graph structure inference framework, called LLM4RGNN, which efficiently utilizes LLMs to make GNNs more robust. Additionally, LLM4RGNN is a general framework, suitable for different LLMs and GNNs.
\item Extensive experiments demonstrate that LLM4RGNN consistently improves the robustness of various GNNs against topology attacks. Even in some cases where the perturbation ratio increases to 40\%, the accuracy of GNNs with LLM4RGNN is still better than that on the clean graph.
\item We utilize GPT-4 to construct an instruction dataset, including GPT-4's maliciousness assessments and analyses of 26,518 edges. This dataset will be publicly released, which can be used to tune any other LLMs so that they can have the robust graph structure inference capability as GPT-4.
\end{itemize}
\section{Preliminaries}
\subsection{Text-attributed Graphs (TAGs)}
Here, a Text-attributed graph (TAG), defined as $\mathcal G=(\mathcal{V},\mathcal{E}, \mathcal{S})$, is a graph with node-level textual information, where $\mathcal{V}=\{v_1,\ldots,v_{\left | \mathcal{V} \right |}\}$, $\mathcal{E}=\{e_1,\ldots,e_{\left | \mathcal{E} \right |}\}$ and $\mathcal{S}=\{s_1,\ldots,s_{\left | \mathcal{V} \right |}\}$ are the node set, edge set, and text set, respectively. The adjacency matrix of the graph $\mathcal G$ is denoted as $\mathbf{A}\in\mathbb{R}^{\left | \mathcal{V} \right | \times \left | \mathcal{V} \right |}$, where $\mathbf{A}_{ij}=1$ if nodes $v_i$ and $v_j$ are connected, otherwise $\mathbf{A}_{ij}=0$. In this work, we focus on the node classification task on TAGs. Specifically, each node $v_i$ corresponds to a label $y_i$ that indicates which category the node $v_i$ belongs to.
Usually, we encode the text set $\mathcal{S}$ as the node feature matrix $\mathbf{X}=\{\mathbf{x}_{1},\dots,\mathbf{x}_{\left | \mathcal{V} \right |}\}$ via some embedding techniques~\cite{mikolov2013distributed, harris1954distributional, chen2024exploring} to train GNNs, where $\mathbf{x}_i \in \mathbb{R}^d$.
Given some labeled nodes $\mathcal{V}_L\subset\mathcal{V}$, the goal is training a GNN $f(\mathbf{A}, \mathbf{X})$ to predict the labels of the remaining unlabeled nodes $\mathcal{V}_U=\mathcal{V}\setminus\mathcal{V}_L$.
\begin{figure}[t]
    \centering
    \includegraphics[width=1\linewidth]{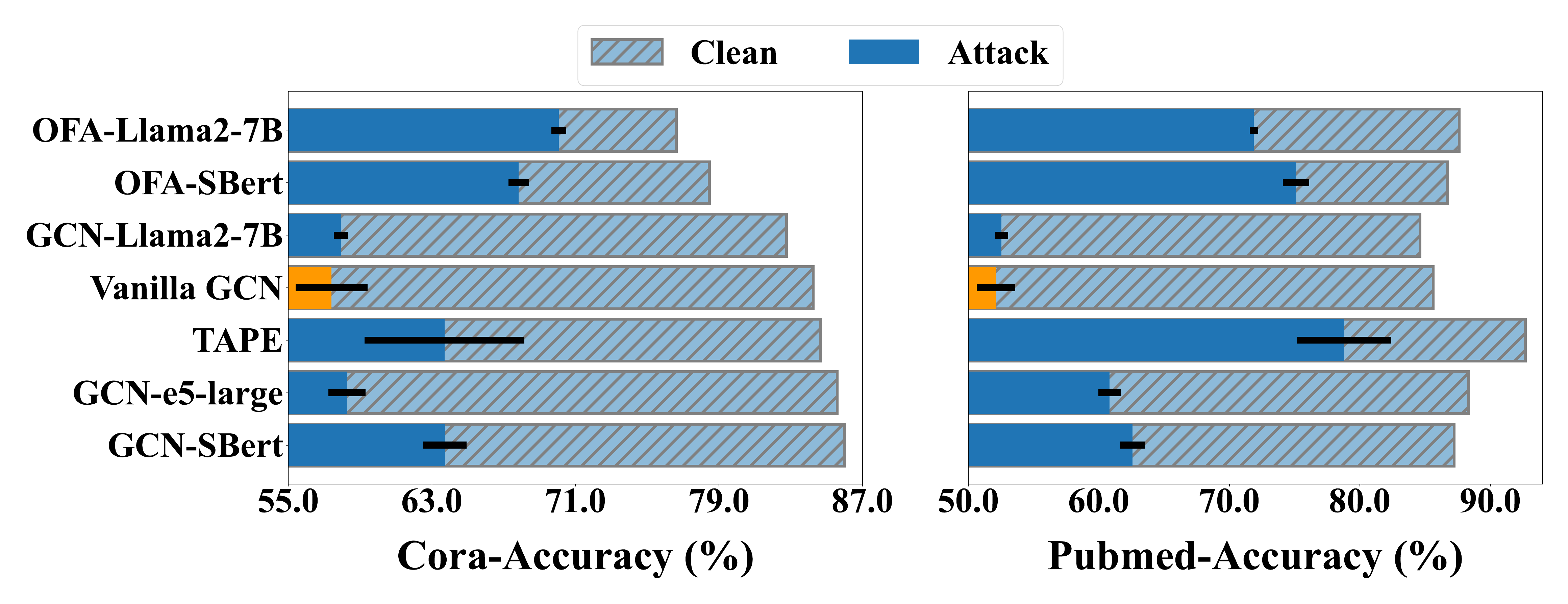}
    \vskip -0.15in
    \caption{The accuracy of different GNNs combining LLMs/LMs against Mettack with a 20\% perturbation rate.}
    \vskip -0.15in
    \label{fig: pre}
\end{figure}
\subsection{Graph Adversarial Robustness}
This paper primarily focuses on stronger poisoning attacks, which can lead to an extremely low model performance by directly modifying the training data~\cite{zhu2019robust, zugner2018adversarial}. The formal definition of adversarial robustness against poisoning attacks is as follows:
\begin{align} \label{eq: def_robustness}
     \max_{\delta \in \Delta} \min_{\theta} \mathcal{L}(f_\theta(\mathcal{G}+\delta), \mathbf{y}_{\mathcal{T}}),
\end{align}
where \( \delta \) represents a perturbation to the graph \( \mathcal{G} \), which may include perturbations to node features, inserting or deleting of edges, etc., \( \Delta \) represents all permitted and effective perturbations. $\mathbf{y}_{\mathcal{T}}$ is the node labels of the target set ${\mathcal{T}}$. $\mathcal{L}$ denotes the training loss of GNNs, and $\theta$ is the model parameters of $f$.
Equation~\ref{eq: def_robustness} indicates that under the worst-case perturbation \( \delta \), the adversarial robustness of model \( f \) is represented by its performance on the target set \( \mathcal{T} \). A smaller loss value suggests stronger adversarial robustness, i.e., better model performance. In this paper, we primarily focus on the robustness under two topology attacks: (1) Targeted attacks~\cite{zugner2018adversarial}, where attackers aim to mislead the model's prediction on speciﬁc nodes $v$ by manipulating the adjacent edges of $v$, thus $\mathcal{T}={v}$. (2) Non-targeted attacks~\cite{zugner_adversarial_2019, waniek2018hiding}, where attackers aim to degrade the overall performance of GNNs but do not care which node is being targeted, thus $\mathcal{T}=V_{\text{test}}$, where $V_{\text{test}}$ denotes the test set.

\begin{figure*}[tp]
    \centering
    \includegraphics[width=1.0\linewidth]{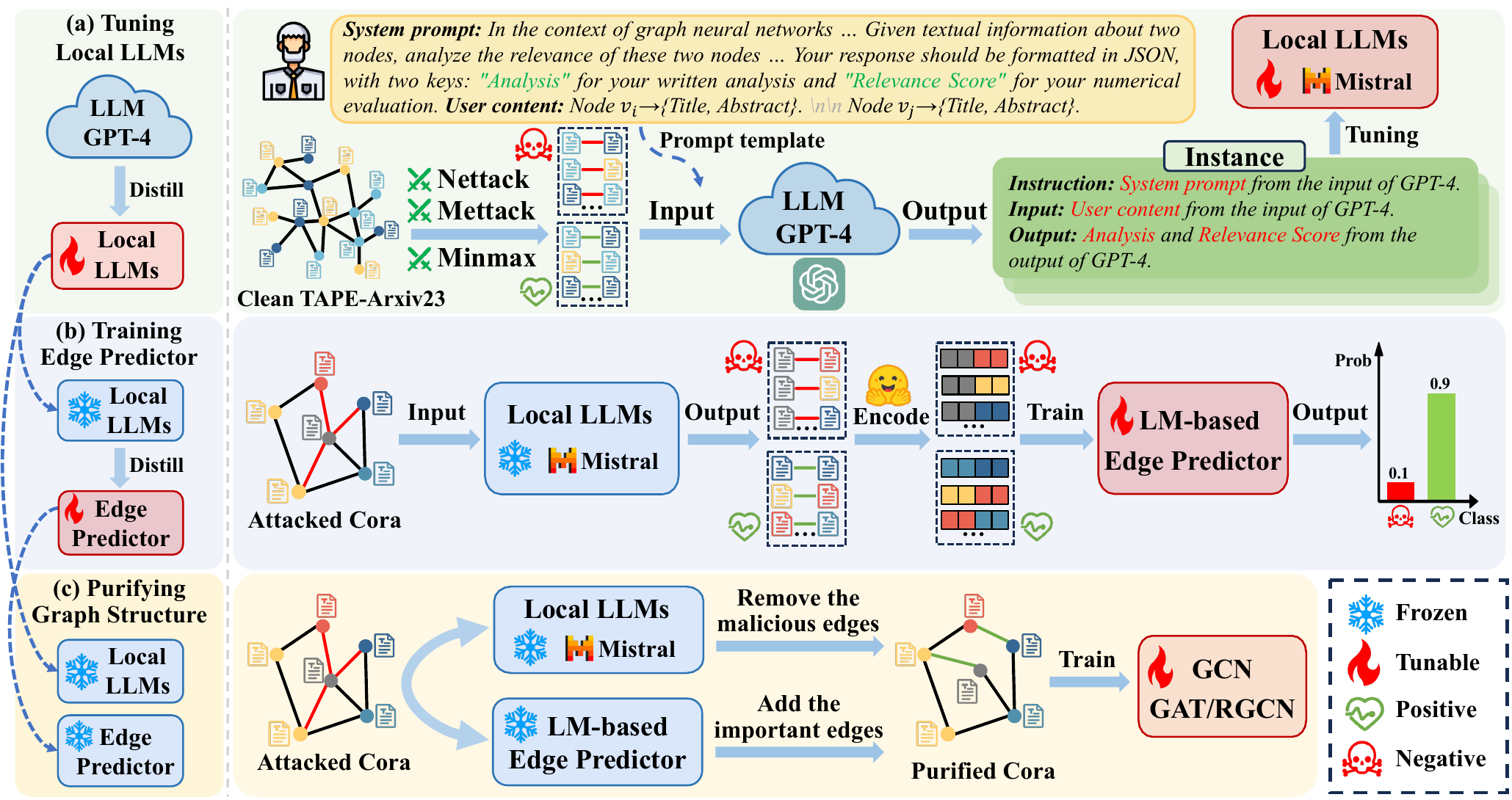}
    \vskip -0.1in
    \caption{Overall framework of LLM4RGNN.
    }
    \vskip -0.175in
    \label{fig: overview}
\end{figure*}
\section{The Adversarial Robustness of GNNs Combining LLMs/LMs} \label{sec: preliminary}
In this section, we empirically investigate whether LLMs enhance or weaken the adversarial robustness of GNNs to a certain extent. Specifically, for the Cora~\cite{mccallum2000automating} and PubMed~\cite{sen2008collective} datasets, based on non-contextualized embeddings encoded by BoW~\cite{harris1954distributional} or TF-IDF~\cite{robertson2004understanding}, we employ Mettack~\cite{zugner_adversarial_2019} with a 20\% perturbation rate to generate attack topology. 
We compare seven representative baselines: TAPE~\cite{he2023harnessing} utilizes LLMs to generate extra semantic knowledge relevant to the nodes. OFA~\cite{liu2023one} employs LLMs to unify different graph data and tasks, where OFA-SBert utilizes Sentence Bert~\cite{reimers2019sentence} to encode the text of nodes, training and testing the GNNs on each dataset independently. OFA-Llama2-7B involves training a single GNN across the Cora, Pubmed, and OGBN-Arxiv~\cite{hu2020open} datasets. Following the work~\cite{chen2024exploring}, GCN-Llama2-7B, GCN-e5-large, and GCN-SBert represent the use of Llama2-7B, e5-large, and Sentence Bert as nodes' text encoders, respectively. The vanilla GCN directly utilizes non-contextual embeddings.
By reporting the node classification accuracy on $V_{\text{test}}$ to evaluate the robustness of models against the Mettack. 
As depicted in Figure~\ref{fig: pre}, we observe that under the influence of Mettack, GNNs combining LLMs/LMs suffer from a maximum accuracy decrease of 37.9\% and an average decrease of 23.1\%, while vanilla GCN~\cite{kipf2016semi} suffers from a maximum accuracy decrease of 39.1\% and an average accuracy decrease of 35.5\%. The results demonstrate that GNNs combining LLMs/LMs remain extremely vulnerable against topology perturbations. Another meaningful observation is that OFA-Llama2-7B, OFA-SBert, and TAPE have significant performance improvements over vanilla GCN. One possible reason is that the introduction of additional knowledge can mitigate the impact of topology perturbations. Specifically, TAPE enhances node features by requesting LLMs to generate additional textual information. OFA incorporates other clean datasets and tasks through LLMs/LMs during the training to mitigate the impact of model poisoning.

\section{LLM4RGNN: The Proposed Framework}
In this section, we propose a novel LLM-based robust graph structure inference framework, LLM4RGNN. 
As shown in Figure~\ref{fig: overview}, LLM4RGNN involves three main parts: (a) instruction tuning a local LLM, which distills the inference capability from GPT-4 into a local LLM for identifying malicious edges; (b) training an LM-based edge predictor, which further distills the inference capability from the local LLM into LM-based edge predictor for finding missing important edges; (c) purifying the graph structure by removing malicious edges and adding important edges, making various GNNs more robust.
\subsection{Instruction Tuning a Local LLM}
Given an attacked graph structure, one straightforward method is to query powerful GPT-4 to identify malicious edges on the graph. However, this method is extremely expensive, because there are \({\left | \mathcal{V} \right |}^2\) different perturbation edges on a graph. For example, for the PubMed~\cite{sen2008collective} dataset with 19,717 nodes, the cost is approximately \$9.72 million. Thus, we hope to distill the inference capability of GPT-4 into a local LLM, to identify malicious edges. To this end, instruction tuning based on GPT-4 is a popular fine-tuning technique~\cite{xu2024survey, chen2023label}, which utilizes GPT-4 to construct an instruction dataset, and then further trains a local LLM in a supervised fashion. The instruction dataset generally consists of instance (instruction, input, output), where instruction denotes the human instruction (a task definition in natural language) for LLMs, input is used as supplementary content for the instruction, and output denotes the desired output that follows the instruction. Therefore, the key is how to construct an effective instruction dataset for fine-tuning an LLM to identify the malicious edges.

In the tuning local LLMs phase of Figure~\ref{fig: overview} (a), based on an open-source and clean graph structure $\mathbf{A}$ (TAPE-Arxiv23~\cite{he2023harnessing}), we utilize the existing attacks (Mettack~\cite{zugner_adversarial_2019}, Nettack~\cite{zugner2018adversarial}, and Minmax~\cite{xu2019topology}) to generate the perturbed graph structure $\mathbf{A}^{\prime}$, thus we have the modification matrix $\mathbf{S}$ as follows:
\begin{align}
    \mathbf{S}=\mathbf{A}-\{\mathbf{A}^{\prime}_{\text{Mettack}}\bigcup\mathbf{A}^{\prime}_{\text{Nettack}}\bigcup\mathbf{A}^{\prime}_{\text{Minmax}}\},
\end{align}
where $\mathbf{S}\in\{-1, 0, 1\}^{{\left | \mathcal{V} \right |}\times {\left | \mathcal{V} \right |}}$ and $\mathbf S_{ij} = \mathbf S_{ji} = -1$ when the edge between nodes $v_i$ and $v_j$ is added. Conversely, it is removed if and only if $\mathbf S_{ij} = \mathbf S_{ji} = 1$, and $\mathbf S_{ij} = \mathbf S_{ji} = 0$ implies that the edge remains unchanged. Here the added edges are considered as negative edge set $\mathcal{E}_{n}$, i.e., malicious edge set, and the removed edges are considered as positive edge set $\mathcal{E}_{p}$, i.e., important edge set. Since the attack methods prefer adding edges over removing edges~\cite{jin2021adversarial}, to balance $\mathcal{E}_{n}$ and $\mathcal{E}_{p}$, we sample a certain number of clean edges from $\mathbf{A}$ to $\mathcal{E}_{p}$. 
With \( \mathcal{E}_{n} \) and \( \mathcal{E}_{p} \), we construct the query edge set $\mathcal{E}_{\text{q}}= \mathcal{E}_{n}\bigcup \mathcal{E}_{p}$, which will be used to construct prompts for requesting GPT-4. 

Next, based on $\mathcal{E}_{q}$, we query the GPT-4 in an open-ended manner. This involves prompting the GPT-4 to make predictions on how malicious an edge is and provide analysis for its decisions. With this objective, we design a prompt template that includes "System prompt", which is an open-ended question about how malicious the edge is, and "User content", which is the textual information of node pairs $\left(v_i, v_j\right)$ from $\mathcal{E}_{p}$. The general structure of the template follows: (where "System prompt" and "User content" also respectively correspond to instruction and input in the instruction dataset.)
\vspace{-2.5pt}
\begin{tcolorbox}[top=2pt, bottom=2pt, left=4pt, right=4pt]
\textbf{System prompt:} In the context of graph neural networks, attackers manipulate models by adding irrelevant edges or removing relevant ones, leading to incorrect predictions. Your role is crucial in defending against such attacks by evaluating the relevance between pairs of nodes, which will help in identifying and removing the irrelevant edges to mitigate the impact of adversarial attacks on graph-based models. Given textual information about two nodes, analyze the relevance of these two nodes. Provide a concise analysis(approximately 100 words) and assign an integer relevance score from 1 to 6, where 1 indicates completely irrelevant and 6 indicates directly relevant. Your response should be formatted in JSON, with two keys: "Analysis" for your written analysis and "Relevance Score" for your numerical evaluation.\\
\textbf{User content:} Node $v_i$$\rightarrow$\{Title, Abstract\}.\mynl\mynl Node $v_j$$\rightarrow$\{Title, Abstract\}.
\end{tcolorbox}
\vspace{-2.5pt}
In the "System prompt", we provide background knowledge about tasks and the specific roles played by LLMs in the prompt, which can more effectively harness the inference capability of GPT-4~\cite{he2023harnessing, yu2023empower}. Additionally, we require GPT-4 to provide a fine-grained rating of the maliciousness of edges on a scale from 1 to 6, where a lower score indicates more malicious, and a higher score indicates more important. The concept of "Analysis" is particularly crucial, as it not only facilitates an inference process in GPT-4 regarding prediction results, but also serves as a key to distilling the inference capability of GPT-4 into local LLMs. Finally, the output of the instruction dataset is generated by GPT-4 as follows:
\vspace{-2.5pt}
\begin{tcolorbox}[top=2pt, bottom=2pt, left=4pt, right=4pt]
\textbf{Analysis:} \{Analysis of predicted results\}. \\
\textbf{Relevance Score:} \{Predicted integer scores from 1-6\}.
\end{tcolorbox}
\vspace{-2.5pt}
In fact, it is difficult for GPT-4 to predict with complete accuracy. To construct a cleaner instruction dataset, we design a post-processing filtering operation. Specifically, for the output of GPT-4, 
we only preserve the edges with relevance scores \( r_e \in \{1, 2, 3\}\) from the negative sample set \(\mathcal{E}_{\text{n}}\), and the edges with \( r_e \in \{4, 5, 6\}\) from the positive sample set \(\mathcal{E}_{\text{q}}\). The refined instruction dataset is then used to fine-tune a local LLM, such as Mistral-7B~\cite{jiang2023mistral} or Llama3-8B~\cite{touvron2023llama}. After that, the well-tuned LLM is able to infer the maliciousness of edges similar to GPT-4. We also provide case studies of GPT-4 and the local LLM (Mistral-7B) in Appendix~\ref{app: case_study}.
\begin{table*}[tbp]
\caption{Node classification accuracy ($\%\pm\sigma$) under non-targeted attack (Mettack). Bolded results indicate improved performance.}
\vskip -0.15in
\label{tab: meta_gnn}
\centering
\resizebox{\linewidth}{!}{
\begin{tabular}{p{0pt}c|c>{\columncolor{myshadow}}c|c>{\columncolor{myshadow}}c|c>{\columncolor{myshadow}}c|c>{\columncolor{myshadow}}c}
\hline
\multirow{2}{*}{} &
\multicolumn{1}{l|}{\multirow{2}{*}{\makecell{Dataset \\ Ptb Rate}}} &
\multicolumn{2}{c|}{GCN} &
\multicolumn{2}{c|}{GAT} &
\multicolumn{2}{c|}{RGCN} &
\multicolumn{2}{c}{SimP-GCN} \\
\cline{3-10}
& \multicolumn{1}{l|}{} &
\multicolumn{1}{c}{Vanilla} &
\multicolumn{1}{c|}{LLM4RGNN} &
\multicolumn{1}{c}{Vanilla} &
\multicolumn{1}{c|}{LLM4RGNN} &
\multicolumn{1}{c}{Vanilla} &
\multicolumn{1}{c|}{LLM4RGNN} &
\multicolumn{1}{c}{Vanilla} &
\multicolumn{1}{c}{LLM4RGNN} \\
\hline
\multirow{4}{*}{\hspace{-10pt} \vspace{-2pt}  \rotcell{\makebox[10pt][c]{Cora}}}      & 0\%                                                   & $84.25_{\pm0.36}$                  & $84.13_{\pm0.33}$                   & ${84.62_{\pm0.47}}$                  & $84.61_{\pm0.39}$                   & $84.61_{\pm0.46}$                  & $84.50_{\pm0.46}$                   & ${84.70_{\pm0.62}}$                  & $84.46_{\pm0.67}$                    \\
                                     & 5\%                                                    & $75.62_{\pm1.70}$                  & $\mathbf{81.76_{\pm0.69}}$                   & $79.82_{\pm0.52}$                  & $\mathbf{81.22_{\pm0.78}}$                   & $76.65_{\pm0.82}$                  & $\mathbf{82.19_{\pm0.59}}$                   & $79.36_{\pm0.99}$                  & $\mathbf{81.78_{\pm0.51}}$                    \\
                                     & 10\%                                                   & $70.72_{\pm2.86}$                  & $\mathbf{81.80_{\pm0.76}}$                   & $75.33_{\pm1.17}$                  & $\mathbf{81.86_{\pm0.61}}$                   & $69.62_{\pm1.09}$                  & $\mathbf{82.14_{\pm0.60}}$                   & $77.00_{\pm1.34}$                  & $\mathbf{82.19_{\pm0.75}}$                    \\
                                     & 20\%                                                   & $57.41_{\pm2.00}$                  & $\mathbf{81.41_{\pm0.77}}$                   & $63.17_{\pm1.82}$                  & $\mathbf{81.00_{\pm0.98}}$                   & $59.27_{\pm0.81}$                  & $\mathbf{81.39_{\pm0.44}}$                   & $76.04_{\pm1.66}$                  & $\mathbf{81.20_{\pm0.63}}$                    \\ 
\hline
\multirow{4}{*}{\hspace{-10pt} \vspace{-3pt}  \rotcell{\makebox[10pt][c]{Citeseer}}}  & 0\%                                                    & $73.38_{\pm0.72}$                  & $\mathbf{74.20_{\pm0.56}}$                   & $73.95_{\pm0.45}$                  & $73.72_{\pm1.07}$                   & ${74.45_{\pm0.31}}$                  & $74.07_{\pm0.90}$                   & $72.71_{\pm1.14}$                  & $\mathbf{73.30_{\pm0.62}}$                    \\
                                     & 5\%                                                    & $69.69_{\pm0.59}$                  & $\mathbf{73.94_{\pm0.56}}$                   & $71.37_{\pm0.73}$                  & $\mathbf{73.22_{\pm1.08}}$                   & $72.30_{\pm0.25}$                  & $\mathbf{73.61_{\pm0.65}}$                   & $71.07_{\pm0.93}$                  & $\mathbf{73.48_{\pm0.58}}$                    \\
                                     & 10\%                                                   & $65.75_{\pm0.94}$                  & $\mathbf{73.62_{\pm0.39}}$                   & $68.40_{\pm1.14}$                  & $\mathbf{73.21_{\pm1.16}}$                   & $69.36_{\pm0.40}$                  & $\mathbf{73.41_{\pm0.90}}$                   & $70.55_{\pm0.55}$                  & $\mathbf{72.84_{\pm0.50}}$                    \\
                                     & 20\%                                                   & $58.72_{\pm1.00}$                  & $\mathbf{74.12_{\pm0.85}}$                   & $61.92_{\pm1.76}$                  & $\mathbf{73.94_{\pm0.67}}$                   & $62.79_{\pm0.60}$                  & $\mathbf{74.04_{\pm0.70}}$                   & $69.54_{\pm0.78}$                  & $\mathbf{73.70_{\pm0.64}}$                    \\ 
\hline
\multirow{4}{*}{\hspace{-10pt} \vspace{-3pt} \rotcell{\makebox[10pt][c]{Pubmed}}}    & 0\%                                                    & $85.62_{\pm0.10}$                  & $\mathbf{86.21_{\pm0.13}}$                   & $85.14_{\pm0.11}$                  & $85.10_{\pm0.09}$                   & $85.84_{\pm0.10}$                  & $\mathbf{86.35_{\pm0.10}}$                   & $87.26_{\pm0.08}$                  & $\mathbf{87.53_{\pm0.14}}$                    \\
                                     & 5\%                                                    & $73.27_{\pm1.15}$                  & $\mathbf{85.27_{\pm0.26}}$                   & $80.87_{\pm0.91}$                  & $\mathbf{84.45_{\pm0.23}}$                   & $81.61_{\pm0.20}$                  & $\mathbf{85.93_{\pm0.10}}$                   & $85.90_{\pm0.20}$                  & $\mathbf{86.08_{\pm0.27}}$                    \\
                                     & 10\%                                                   & $67.54_{\pm0.42}$                 & $\mathbf{84.95_{\pm0.17}}$                    & $69.94_{\pm4.60}$                  & $\mathbf{84.88_{\pm0.19}}$                   & $69.55_{\pm0.40}$                  & $\mathbf{85.79_{\pm0.22}}$                   & $85.71_{\pm0.11}$                  & $\mathbf{86.06_{\pm0.13}}$                    \\
                                     & 20\%                                                   & $52.12_{\pm1.47}$                  & $\mathbf{84.99_{\pm0.32}}$                   & $53.07_{\pm0.69}$                  & $\mathbf{85.20_{\pm0.30}}$                   & $48.47_{\pm0.71}$                  & $\mathbf{85.99_{\pm0.21}}$                   & $85.70_{\pm0.15}$                  & $\mathbf{86.07_{\pm0.24}}$                    \\ 
\hline
\multirow{4}{*}{\hspace{-10pt} \vspace{-2.5pt} \rotcell{\makebox[10pt][c]{Arxiv}}} & 0\%                                                    & $67.01_{\pm0.08}$                  & $\mathbf{68.16_{\pm0.36}}$                   & $65.02_{\pm0.37}$                  & $\mathbf{67.43_{\pm0.49}}$                   & $65.53_{\pm0.38}$                  & $\mathbf{67.53_{\pm0.38}}$                   & $65.13_{\pm0.31}$                  & $\mathbf{66.20_{\pm1.16}}$                    \\
& 5\%                                                    & $50.51_{\pm0.29}$                  & $\mathbf{68.86_{\pm0.41}}$                   & $54.68_{\pm0.57}$                  & $\mathbf{68.46_{\pm0.47}}$                   & $52.35_{\pm0.09}$                  & $\mathbf{68.71_{\pm0.43}}$                   & $51.07_{\pm3.87}$                  & $\mathbf{67.88_{\pm0.47}}$                    \\
& 10\%                                                   & $42.91_{\pm0.81}$                  & $\mathbf{68.65_{\pm0.43}}$                   & $49.18_{\pm0.54}$                  & $\mathbf{68.56_{\pm0.55}}$                   & $44.75_{\pm0.36}$                  & $\mathbf{68.54_{\pm0.70}}$                   & $46.52_{\pm6.76}$                  & $\mathbf{66.47_{\pm2.54}}$                    \\
& 20\%                                                   & $33.96_{\pm0.46}$                  & $\mathbf{69.17_{\pm0.43}}$                   & $34.24_{\pm2.12}$                  & $\mathbf{68.86_{\pm0.54}}$                   & $31.69_{\pm0.63}$                  & $\mathbf{69.00_{\pm0.59}}$                   & $37.31_{\pm7.59}$                  & $\mathbf{68.15_{\pm0.65}}$                    \\
\hline
\multirow{4}{*}{\hspace{-10pt} \vspace{-3pt} \rotcell{\makebox[10pt][c]{Products}}} 
& 0\% & $79.86_{\pm0.15}$ & $79.04_{\pm0.42}$ & $78.75_{\pm0.26}$ & $77.76_{\pm0.62}$ & $78.45_{\pm0.25}$ & $77.83_{\pm0.48}$ & $75.87_{\pm0.36}$ & $75.84_{\pm0.29}$ \\
& 5\% & $66.62_{\pm0.56}$ & $\mathbf{76.34_{\pm0.39}}$ & $76.41_{\pm0.60}$ & $\mathbf{76.70_{\pm0.68}}$ & $71.48_{\pm0.35}$ & $\mathbf{75.20_{\pm0.25}}$ & $64.84_{\pm0.60}$ & $\mathbf{72.67_{\pm0.66}}$ \\
& 10\% & $63.31_{\pm0.52}$ & $\mathbf{75.80_{\pm0.24}}$ & $74.13_{\pm0.37}$ & $\mathbf{75.48_{\pm0.75}}$ & $68.98_{\pm0.41}$ & $\mathbf{74.75_{\pm0.38}}$ & $58.59_{\pm1.97}$ & $\mathbf{71.80_{\pm0.48}}$ \\
& 20\% & $57.56_{\pm0.64}$ & $\mathbf{76.57_{\pm0.46}}$ & $70.25_{\pm0.82}$ & $\mathbf{74.98_{\pm0.53}}$ & $64.81_{\pm0.31}$ & $\mathbf{74.32_{\pm0.38}}$ & $50.36_{\pm0.96}$ & $\mathbf{73.71_{\pm0.81}}$ \\
\hline
\end{tabular}}
\vskip -0.15in
\end{table*}

\subsection{Training an LM-based Edge Predictor}
Now, given a new attacked graph structure \( \mathbf{A}^{\prime} \), our key idea is to recover a robust graph structure $\hat{\mathbf{A}}$. Intuitively, we can input each edge of \( \mathbf{A}^{\prime} \) into the local LLM and obtain its relevance score \( r_e \). By removing edges with lower scores, we can mitigate the impact of malicious edges on model predictions. Meanwhile, considering that attackers can also delete some important edges to reduce model performance, we need to find and add important edges that do not exist in \( \mathbf{A}^{\prime} \). Although the local LLM can identify important edges with higher relevance scores, it is still very time and resource-consuming with \({\left | \mathcal{V} \right |}^2\) edges. Therefore, we further design an LM-based edge predictor, as depicted in Figure~\ref{fig: overview} (b), which utilizes Sentence Bert~\cite{reimers2019sentence} as the text encoder and trains a lightweight multilayer perceptron (MLP) to find missing important edges.

Firstly, we introduce how to construct the feature of each edge. Inspired by~\cite{chen2024exploring}, deep sentence embeddings have emerged as a powerful text encoding method, outperforming non-contextualized embeddings~\cite{harris1954distributional, robertson2004understanding}. Furthermore, sentence embedding models offer a lightweight method to obtain representations without fine-tuning. Consequently, for each node $v_i$, we adopt a sentence embedding model $\text{LM}$ as texts encoder to extract representations $\mathbf{h}_i$ from the raw text $s_{i}$, i.e., $\mathbf{h}_i=\mathrm{\text{LM}}(s_{i})$. We concatenate the representations of the node $v_i$ and $v_j$ as the feature for the corresponding edge.

Then the edge label $y_e$ can be derived from $r_e$ as follows:
\begin{align}
    y_e=\begin{cases}1&\text{if} \ r_e > 4 \\0&\text{if} \ r_e \leq 4 \end{cases},
\end{align}
here, we utilize the local LLM as an edge annotator to distill its inference capability, and select $4$ as the threshold to find the most positive edges. It is noted that there may be a label imbalance problem, where the number of positive edges is much higher than the negative. Thus, based on the cosine similarity, we select some node pairs with a lower similarity to construct a candidate set. When there are not enough negative edges, we sample from the candidate set to balance the training set. 

Next, we feed the feature of each edge into an $\text{MLP}$ to obtain the prediction probability $\hat{y}_e(v_i, v_j)=\text{MLP}(\mathbf{h}_i||\mathbf{h}_j)$.
 The cross-entropy loss function is used to optimize the parameters of $\text{MLP}$ as:
 \begin{align}
     \mathcal{L}_{\mathrm{CE}}(y_e,\hat{y}_e)=-[y_e\log(\hat{y}_e)+(1-y_e)\log(1-\hat{y}_e)].
 \end{align}
 After training the edge predictor, we input any node pair ($v_i$, $v_j$) that does not exist in $\mathbf{A}^{\prime}$ into it to obtain the prediction probability of edge existence. We have the important edge set for node $v_i$:
\begin{align}
    \mathcal{E}_{v_i} = \{ (v_i, v_j) \mid \ j\neq i, \ \mathbf{A}^{\prime}_{ij}=0, \ 
    \hat{y}_e > \gamma \ &\text{and} \ \hat{y}_e \in \text{Top}_K \},
\end{align}
where $\gamma \in (0, 1)$ is the threshold and $K$ is the maximum number of edges. In this way,
we can select the top $K$ neighbors for the current node $v_i$ with predicted score greater than threshold $\gamma$, to establish the most important edges for $v_i$ as possible. 
For all the nodes, we have the final important edge set $\mathcal{E}_{\text{add}} = \bigcup_{v_i \in \mathcal{V}}\mathcal{E}_{v_i}$.

\subsection{Purifying Attacked Graph Structure}
\vspace{-2.5pt}
In Figure~\ref{fig: overview} (c), the robust graph structure \( \hat{\mathbf{A}} \) is derived from the purification of \( \mathbf{A}^{\prime} \). Specifically, new edges from \( \mathcal{E}_{\text{add}} \) will be added in \( \mathbf{A}^{\prime} \). Simultaneously, with the relevance score \( r_e \) of each edge, we remove the malicious edges in \( \mathbf{A}^{\prime} \) by setting a purification threshold \( \beta \), i.e., edges with \( r_e \) larger than \( \beta \) are preserved, otherwise removed. The $\hat{\mathbf{A}}$ is adaptive to any GNNs, making GNNs more robust.

\vspace{-2.5pt}
\section{Experiments}\label{sec: exp}
\vspace{-2.5pt}
\subsection{Experimental Setup}
\vspace{-1.5pt}
\textbf{Dataset.} We conduct experiments on four cross-dataset citation networks (Cora \cite{mccallum2000automating}, Citeseer \cite{giles1998citeseer}, Pubmed \cite{sen2008collective}, OGBN-Arxiv \cite{hu2020open}) and one cross-domain product network (OGBN-Products \cite{hu2020open}). We report the average performance and standard deviation over 10 seeds for each result. More details refer to Appendix~\ref{sec: exp-dataset}.\\
\textbf{Baseline.} First, LLM4RGNN is a general LLM-based framework to enhance the robustness of GNNs. Therefore, we select the classical GCN~\cite{kipf2016semi} and three robust GNNs (GAT~\cite{velivckovic2017graph}, RGCN~\cite{zhu2019robust} and Simp-GCN~\cite{jin2021node}) as baselines. Moreover, to more comprehensively evaluate LLM4RGNN, we also compare it with existing SOTA robust GNN frameworks, including ProGNN~\cite{jin2020graph}, STABLE~\cite{li2022reliable}, HANG-quad~\cite{zhao2024adversarial} and GraphEdit\footnote{GraphEdit only provides prompts for Cora, Citeseer and Pubmed datasets.}~\cite{guo2024graphedit}, where GCN is selected as the object for improving robustness. More baseline introduction and implementation details refer to Appendix~\ref{sec:exp-baseline} and ~\ref{sec:exp-implementation}, respectively.
\vspace{-5pt}
\subsection{Main Result}
\vspace{-1pt}
In this subsection, we conduct extensive evaluations of LLM4RGNN against three popular poisoning topology attacks: non-targeted attacks Mettack~\cite{zugner_adversarial_2019} and DICE~\cite{waniek2018hiding}, and targeted attack Nettack~\cite{zugner2018adversarial}. We report the accuracy (ACC (↑)) on representative transductive node classification task. More results of inductive poisoning attacks refer to Appendix~\ref{exp: ind_poison}.
\begin{table*}[htp]
\caption{Node classification accuracy ($\%\pm\sigma$) under non-targeted attack (DICE). Bolded results indicate improved performance.} \label{tab: dice_gnn}
\vskip -0.15in
\resizebox{\linewidth}{!}{
\centering
\begin{tabular}{p{0pt}c|c>{\columncolor{myshadow}}c|c>{\columncolor{myshadow}}c|c>{\columncolor{myshadow}}c|c>{\columncolor{myshadow}}c} 
\hline
\multirow{2}{*}{} &
\multicolumn{1}{c|}{\multirow{2}{*}{\makecell{Dataset \\ Ptb Rate}}} &
\multicolumn{2}{c|}{GCN} &
\multicolumn{2}{c|}{GAT} &
\multicolumn{2}{c|}{RGCN} &
\multicolumn{2}{c}{SimP-GCN} \\
\cline{3-10}
& \multicolumn{1}{c|}{} &
\multicolumn{1}{c}{Vanilla} &
\multicolumn{1}{c|}{LLM4RGNN} &
\multicolumn{1}{c}{Vanilla} &
\multicolumn{1}{c|}{LLM4RGNN} &
\multicolumn{1}{c}{Vanilla} &
\multicolumn{1}{c|}{LLM4RGNN} &
\multicolumn{1}{c}{Vanilla} &
\multicolumn{1}{c}{LLM4RGNN} \\
\hline
\multirow{4}{*}{\hspace{-10pt} \vspace{10pt} \rotcell{\makebox[7.5pt][c]{ Cora}}}      
& 10\%   & $81.38_{\pm 0.31}$  & $\mathbf{81.84_{\pm 0.71}}$  & $81.52_{\pm 0.49}$  & $\mathbf{82.12_{\pm 0.71}}$  & $81.71_{\pm 0.52}$  & $\mathbf{82.14_{\pm 0.52}}$  & $82.29_{\pm 0.55}$  & $\mathbf{82.35_{\pm 0.65}}$  \\
& 20\%   & $78.38_{\pm 0.51}$  & $\mathbf{80.96_{\pm 0.61}}$  & $78.51_{\pm 0.35}$  & $\mathbf{81.02_{\pm 1.00}}$  & $78.58_{\pm 0.55}$  & $\mathbf{81.52_{\pm 0.88}}$  & $79.65_{\pm 0.61}$  & $\mathbf{81.38_{\pm 0.56}}$  \\
& 40\%   & $71.62_{\pm 0.40}$  & $\mathbf{79.46_{\pm 0.93}}$  & $71.81_{\pm 0.72}$  & $\mathbf{79.11_{\pm 1.00}}$  & $72.77_{\pm 0.48}$  & $\mathbf{79.86_{\pm 0.86}}$  & $74.37_{\pm 0.67}$  & $\mathbf{79.64_{\pm 1.14}}$  \\ 
\hline
\multirow{4}{*}{\hspace{-10pt} \vspace{11pt} \rotcell{\makebox[7.5pt][c]{Citeseer}}}  
& 10\%   & $71.08_{\pm 0.63}$  & $\mathbf{73.95_{\pm 0.59}}$  & $71.62_{\pm 0.79}$  & $\mathbf{73.73_{\pm 0.58}}$  & $72.74_{\pm 0.46}$  & $\mathbf{73.85_{\pm 1.07}}$  & $71.15_{\pm 0.75}$  & $\mathbf{73.41_{\pm 0.64}}$  \\
& 20\%   & $68.79_{\pm 0.49}$  & $\mathbf{73.32_{\pm 0.58}}$  & $69.68_{\pm 0.59}$  & $\mathbf{73.26_{\pm 0.45}}$  & $70.67_{\pm 0.33}$  & $\mathbf{73.61_{\pm 0.69}}$  & $70.67_{\pm 0.53}$  & $\mathbf{72.66_{\pm 0.73}}$  \\
& 40\%   & $63.98_{\pm 1.02}$  & $\mathbf{73.35_{\pm 0.50}}$  & $64.81_{\pm 0.87}$  & $\mathbf{73.07_{\pm 0.71}}$  & $65.46_{\pm 0.55}$  & $\mathbf{73.64_{\pm 1.02}}$  & $68.53_{\pm 1.80}$  & $\mathbf{72.80_{\pm 0.63}}$  \\ 
\hline
\multirow{4}{*}{\hspace{-10pt} \vspace{11.5pt} \rotcell{\makebox[7.5pt][c]{Pubmed}}}    
& 10\%   & $82.86_{\pm 0.17}$  & $\mathbf{85.14_{\pm 0.22}}$  & $81.85_{\pm 0.17}$  & $\mathbf{83.96_{\pm 0.22}}$  & $83.48_{\pm 0.12}$  & $\mathbf{85.24_{\pm 0.15}}$  & ${86.60_{\pm 0.07}}$  & $\mathbf{86.92_{\pm 0.13}}$  \\
& 20\%   & $80.11_{\pm 0.16}$  & $\mathbf{84.96_{\pm 0.19}}$  & $78.92_{\pm 0.20}$  & $\mathbf{84.33_{\pm 0.28}}$  & $80.55_{\pm 0.11}$  & $\mathbf{85.06_{\pm 0.20}}$  & ${85.98_{\pm 0.08}}$  & $\mathbf{86.66_{\pm 0.18}}$  \\
& 40\%   & $74.45_{\pm 0.19}$  & $\mathbf{84.89_{\pm 0.21}}$  & $72.47_{\pm 0.24}$  & $\mathbf{83.98_{\pm 0.17}}$  & $74.78_{\pm 0.18}$  & $\mathbf{84.73_{\pm 0.25}}$  & ${85.36_{\pm 0.09}}$  & $\mathbf{86.42_{\pm 0.11}}$  \\ 
\hline
\multirow{4}{*}{\hspace{-10pt} \vspace{12pt} \rotcell{\makebox[7.5pt][c]{Arxiv}}}
& 10\%   & $64.23_{\pm 0.10}$  & $\mathbf{68.15_{\pm 0.89}}$  & $62.39_{\pm 0.28}$  & $\mathbf{67.96_{\pm 1.05}}$  & $62.92_{\pm 0.27}$  & $\mathbf{67.99_{\pm 1.00}}$  & $62.73_{\pm 0.27}$  & $\mathbf{66.77_{\pm 1.51}}$  \\
& 20\%   & $62.35_{\pm 0.17}$  & $\mathbf{68.21_{\pm 1.01}}$  & $59.60_{\pm 0.40}$  & $\mathbf{67.84_{\pm 1.10}}$  & $60.58_{\pm 0.27}$  & $\mathbf{67.95_{\pm 1.11}}$  & $60.52_{\pm 0.25}$  & $\mathbf{66.56_{\pm 1.19}}$  \\
& 40\%   & $57.55_{\pm 0.14}$  & $\mathbf{68.73_{\pm 0.32}}$  & $54.86_{\pm 0.38}$  & $\mathbf{68.52_{\pm 0.53}}$  & $56.33_{\pm 0.21}$  & $\mathbf{68.47_{\pm 0.40}}$  & $56.38_{\pm 0.23}$  & $\mathbf{67.37_{\pm 1.64}}$  \\
\hline
\multirow{4}{*}{\hspace{-10pt} \vspace{11pt} \rotcell{\makebox[7.5pt][c]{Products}}}
& 10\% & $76.05_{\pm0.16}$ & $\mathbf{77.66_{\pm0.22}}$ & $75.12_{\pm0.28}$ & $\mathbf{76.16_{\pm0.31}}$ & $74.78_{\pm0.33}$ & $\mathbf{76.41_{\pm0.43}}$ & $71.62_{\pm0.61}$ & $\mathbf{74.20_{\pm0.41}}$ \\
& 20\% & $72.48_{\pm0.21}$ & $\mathbf{77.15_{\pm0.62}}$ & $72.03_{\pm0.37}$ & $\mathbf{75.62_{\pm1.06}}$ & $71.71_{\pm0.27}$ & $\mathbf{76.12_{\pm0.59}}$ & $68.03_{\pm0.44}$ & $\mathbf{74.57_{\pm1.00}}$ \\
& 40\% & $66.26_{\pm0.37}$ & $\mathbf{77.02_{\pm0.56}}$ & $65.52_{\pm0.62}$ & $\mathbf{76.20_{\pm0.40}}$ & $64.70_{\pm0.31}$ & $\mathbf{75.85_{\pm0.46}}$ & $60.41_{\pm0.57}$ & $\mathbf{75.01_{\pm0.64}}$ \\
\hline
\end{tabular}}
\vskip -0.15in
\end{table*}

\subsubsection{\textbf{Against Mettack.}}
Non-targeted attacks aim to disrupt the entire graph topology to degrade the performance of GNNs on the test set. We employ the SOTA non-targeted attack method, Mettack~\cite{zugner_adversarial_2019}, and vary the perturbation rate from 0 to 20\% with a step of 5\%. We have the following observations:
(1) From Table~\ref{tab: meta_gnn}, LLM4RGNN consistently improves the robustness across various GNNs. For GCN, there is an average accuracy improvement of 24.3\% and a maximum improvement of 103\% across five datasets. For robust GNNs, including GAT, RGCN and Simp-GCN, LLM4RGNN on average has 16.6\%, 21.4\%, and 13.7\% relative improvements in accuracy. 
Notably, despite fine-tuning the local LLM on the TAPE-Arxiv23 dataset, which does not include any medical or product samples, there is still a relative accuracy improvement of 18.8\% and 11.4\% on the Pubmed and OGBN-Products, respectively.
(2) Referring to Table~\ref{tab: meta_framework}, compared with existing robust GNN frameworks, LLM4RGNN achieves SOTA robustness, which benefits from the powerful understanding and inference capabilities of LLMs.
(3) Combining Table~\ref{tab: meta_gnn} and Table~\ref{tab: meta_framework},
even in some cases where the perturbation ratio increases to 20\%, after using LLM4RGNN to purify the graph structure, the accuracy of GNNs is better than that on the clean graph. One possible reason is that the local LLM effectively identifies malicious edges as negative samples, which helps train a more effective edge predictor to find missing important edges.
\begin{table}[htb]
\caption{Node classification accuracy (\%$\pm\sigma$) under non-targeted attack (Mettack). The best results are in bold. OOT means that the result could not be obtained in 15 days.}
\vskip -0.15in
\label{tab: meta_framework}
\resizebox{\linewidth}{!}{
\centering
\Huge
\begin{tabular}{p{0cm}c|cccc>{\columncolor{myshadow}}c}
\hline
\multirow{2}{*}{} &
\multicolumn{1}{c|}{\makecell{Dataset \\ Ptb Rate}} &
\multicolumn{1}{c}{Pro-GNN} &
\multicolumn{1}{c}{STABLE} &
\multicolumn{1}{c}{HANG-quad} &
\multicolumn{1}{c}{GraphEdit} &
\multicolumn{1}{c}{LLM4RGNN} \\
\hline
\multirow{4}{*}{ \hspace{-25pt} \rotcell{\makebox[12pt][c]{Cora}}}      & 0\%  & \multicolumn{1}{l}{$80.85_{\pm0.44}$}  & $\mathbf{85.09_{\pm0.21}}$  & $79.41_{\pm0.55}$  & $76.35_{\pm0.38}$  & $84.13_{\pm0.33}$  \\
                                     & 5\%  & \multicolumn{1}{l}{$79.81_{\pm0.45}$}  & $80.53_{\pm1.13}$  & $78.81_{\pm1.02}$  & $74.97_{\pm0.95}$  & $\mathbf{81.76_{\pm0.69}}$  \\
                                     & 10\% & \multicolumn{1}{l}{$78.57_{\pm0.97}$}  & $79.53_{\pm0.52}$  & $78.15_{\pm1.14}$  & $74.87_{\pm0.75}$  & $\mathbf{81.80_{\pm0.76}}$  \\
                                     & 20\% & \multicolumn{1}{l}{$76.07_{\pm0.57}$}  & $78.70_{\pm0.89}$  & $74.90_{\pm0.65}$  & $73.82_{\pm0.48}$  & $\mathbf{81.41_{\pm0.77}}$  \\ 
\hline
\multirow{4}{*}{ \hspace{-25pt} \rotcell{\makebox[12pt][c]{Citeseer}}}  & 0\%  & \multicolumn{1}{l}{$71.11_{\pm0.45}$}  & $72.51_{\pm1.37}$  & $71.18_{\pm0.64}$  & $72.60_{\pm0.57}$  & $\mathbf{74.20_{\pm0.56}}$  \\
                                     & 5\%  & \multicolumn{1}{l}{$69.68_{\pm0.53}$}  & $71.23_{\pm1.35}$  & $71.16_{\pm0.71}$  & $71.40_{\pm0.72}$  & $\mathbf{73.94_{\pm0.56}}$  \\
                                     & 10\% & \multicolumn{1}{l}{$68.73_{\pm0.79}$}  & $70.47_{\pm1.47}$  & $70.84_{\pm1.00}$  & $71.35_{\pm0.98}$  & $\mathbf{73.62_{\pm0.39}}$  \\
                                     & 20\% & \multicolumn{1}{l}{$68.37_{\pm0.84}$}  & $67.91_{\pm1.98}$  & $70.00_{\pm1.11}$  & $69.44_{\pm0.86}$  & $\mathbf{74.12_{\pm0.85}}$  \\ 
\hline
\multirow{4}{*}{ \hspace{-25pt} \rotcell{\makebox[12pt][c]{Pubmed}}}    & 0\%  & \multicolumn{1}{c}{OOT}  & $84.07_{\pm0.17}$  & $84.98_{\pm0.13}$  & $85.38_{\pm0.07}$  & $\mathbf{86.21_{\pm0.13}}$  \\
                                     & 5\%  & \multicolumn{1}{c}{OOT}  & $79.41_{\pm0.86}$  & $84.97_{\pm0.16}$  & $84.77_{\pm0.09}$  & $\mathbf{85.27_{\pm0.26}}$  \\
                                     & 10\% & \multicolumn{1}{c}{OOT}  & $77.65_{\pm0.25}$  & $84.88_{\pm0.19}$  & $83.41_{\pm0.19}$  & $\mathbf{84.95_{\pm0.17}}$  \\
                                     & 20\% & \multicolumn{1}{c}{OOT}  & $72.51_{\pm1.05}$  & $84.94_{\pm0.19}$  & $82.34_{\pm0.27}$  & $\mathbf{84.99_{\pm0.32}}$  \\ 
\hline
\multirow{4}{*}{ \hspace{-25pt} \rotcell{\makebox[12pt][c]{Arxiv}}}     & 0\%  & \multicolumn{1}{c}{OOT}  & $66.70_{\pm0.19}$  & $67.85_{\pm0.17}$  & -  & $\mathbf{68.16_{\pm0.36}}$  \\
                                     & 5\%  & \multicolumn{1}{c}{OOT}  & $61.40_{\pm0.50}$  & $63.53_{\pm0.42}$  & -  & $\mathbf{68.86_{\pm0.41}}$  \\
                                     & 10\% & \multicolumn{1}{c}{OOT}  & $59.37_{\pm0.40}$  & $57.49_{\pm0.76}$  & -  & $\mathbf{68.65_{\pm0.43}}$  \\
                                     & 20\% & \multicolumn{1}{c}{OOT}  & $58.24_{\pm0.60}$  & $41.61_{\pm1.28}$  & - & $\mathbf{69.17_{\pm0.43}}$  \\
\hline
\multirow{4}{*}{ \hspace{-25pt} \rotcell{\makebox[12pt][c]{Products}}}     & 0\%  & \multicolumn{1}{c}{OOT}  & $79.23_{\pm0.33}$  & $\mathbf{80.61_{\pm0.18}}$  & -  & $79.04_{\pm0.42}$  \\
                                     & 5\%  & \multicolumn{1}{c}{OOT}  & $76.24_{\pm0.44}$  & $72.41_{\pm0.66}$  & -  & $\mathbf{76.34_{\pm0.39}}$  \\
                                     & 10\% & \multicolumn{1}{c}{OOT}  & $74.38_{\pm0.40}$  & $69.88_{\pm0.86}$  & -  & $\mathbf{75.80_{\pm0.24}}$  \\
                                     & 20\% & \multicolumn{1}{c}{OOT}  & $72.21_{\pm0.40}$  & $62.24_{\pm0.28}$  & - & $\mathbf{76.57_{\pm0.46}}$  \\
\hline
\end{tabular}}
\vskip -0.2in
\end{table}

\begin{table}[htp]
\caption{Node classification accuracy ($\%\pm\sigma$) under non-targeted attack (DICE). The best results are in bold. OOT means that the result could not be obtained in 15 days.}
\vskip -0.15in
\label{tab: dice_framework}
\Huge
\resizebox{\linewidth}{!}{
\centering
\begin{tabular}{p{0cm}c|cccc>{\columncolor{myshadow}}c}
\hline
\multirow{2}{*}{} &
\multicolumn{1}{c|}{\makecell{Dataset \\ Ptb Rate}} &
\multicolumn{1}{c}{Pro-GNN} &
\multicolumn{1}{c}{STABLE} &
\multicolumn{1}{c}{HANG-quad} &
\multicolumn{1}{c}{GraphEdit} &
\multicolumn{1}{c}{LLM4RGNN} \\
\hline
\multirow{4}{*}{ \hspace{-25pt} \rotcell{\makebox[12pt][c]{Cora}}}      & 0\%                                            & $80.85_{\pm 0.44}$ & $\mathbf{85.09_{\pm 0.21}}$ & $79.41_{\pm 0.55}$  & $76.35_{\pm 0.38}$ & $84.13_{\pm 0.33}$\\
                                     & 10\%                                           & $81.56_{\pm 0.36}$ & $81.33_{\pm 0.72}$ & $78.26_{\pm 0.29}$ & $73.73_{\pm 0.57}$ & $\mathbf{81.84_{\pm 0.71}}$  \\
                                     & 20\%                                           & $78.32_{\pm 0.33}$ & $78.80_{\pm 0.83}$ & $76.45_{\pm 0.50}$ & $69.86_{\pm 0.71}$ & $\mathbf{80.96_{\pm 0.61}}$  \\
                                     & 40\%                                           & $71.76_{\pm 0.34}$ & $76.72_{\pm 0.93}$ & $74.28_{\pm 0.42}$ & $66.95_{\pm 0.48}$ & $\mathbf{79.46_{\pm 0.93}}$  \\ 
\hline
\multirow{4}{*}{ \hspace{-25pt} \rotcell{\makebox[12pt][c]{Citeseer}}}  & 0\%                                            & $71.11_{\pm 0.45}$ & $72.51_{\pm 1.37}$ & $71.18_{\pm 0.64}$  & $72.60_{\pm 0.57}$ & $\mathbf{74.20_{\pm 0.56}}$ \\
                                     & 10\%                                           & $70.81_{\pm 0.68}$ & $70.27_{\pm 1.62}$ & $70.29_{\pm 0.70}$ & $68.99_{\pm 0.52}$ & $\mathbf{73.95_{\pm 0.59}}$  \\
                                     & 20\%                                           & $69.21_{\pm 0.79}$ & $69.11_{\pm 1.57}$ & $70.05_{\pm 0.77}$ & $69.31_{\pm 0.41}$ & $\mathbf{73.32_{\pm 0.58}}$ \\
                                     & 40\%                                           & $68.04_{\pm 1.38}$ & $67.55_{\pm 0.86}$ & $68.79_{\pm 1.09}$ & $66.59_{\pm 0.74}$ & $\mathbf{73.35_{\pm 0.50}}$  \\ 
\hline
\multirow{4}{*}{ \hspace{-25pt} \rotcell{\makebox[12pt][c]{Pubmed}}}    & 0\%                                            & OOT                & $84.07_{\pm 0.17}$ & $84.98_{\pm 0.13}$ & $85.38_{\pm 0.07}$ & $\mathbf{86.21_{\pm 0.13}}$  \\
                                     & 10\%                                           & OOT                & $81.68_{\pm 0.21}$ & $84.97_{\pm 0.15}$ & $82.12_{\pm 0.07}$ & $\mathbf{85.14_{\pm 0.22}}$  \\
                                     & 20\%                                           & OOT                & $79.08_{\pm 0.23}$ & $84.94_{\pm 0.18}$  & $81.11_{\pm 0.10}$ & $\mathbf{84.96_{\pm 0.19}}$ \\
                                     & 40\%                                           & OOT                & $74.51_{\pm 0.45}$ & $84.86_{\pm 0.20}$ & $83.04_{\pm 0.09}$ & $\mathbf{84.89_{\pm 0.21}}$ \\ 
\hline
\multirow{4}{*}{ \hspace{-25pt} \rotcell{\makebox[12pt][c]{Arxiv}}}
                                     & 0\%                                            & OOT                & $66.70_{\pm 0.19}$ & $67.85_{\pm 0.17}$& - & $\mathbf{68.16_{\pm 0.36}}$  \\
                                     & 10\%                                           & OOT                & $64.93_{\pm 0.19}$ & $65.99_{\pm 0.19}$ & -& $\mathbf{68.15_{\pm 0.89}}$  \\
                                     & 20\%                                           & OOT                & $62.64_{\pm 0.44}$ & $63.74_{\pm 0.23}$ & - & $\mathbf{68.21_{\pm 1.01}}$  \\
                                     & 40\%                                           & OOT                & $58.55_{\pm 0.37}$ & $59.23_{\pm 0.17}$  & - & $\mathbf{68.73_{\pm 0.32}}$ \\
\hline
\multirow{4}{*}{ \hspace{-25pt} \rotcell{\makebox[12pt][c]{Products}}}
    & 0\%                                            & OOT                & $79.23_{\pm 0.33}$ & $\mathbf{80.61_{\pm 0.18}}$& - & $79.04_{\pm 0.42}$  \\
    & 10\% & \multicolumn{1}{c}{OOT}  & $76.72_{\pm0.32}$  & $77.06_{\pm0.29}$  & -  & $\mathbf{77.66_{\pm0.22}}$  \\
                                     & 20\% & \multicolumn{1}{c}{OOT}  & $75.15_{\pm0.27}$  & $75.52_{\pm0.24}$  & -  & $\mathbf{77.15_{\pm0.62}}$  \\
                                     & 40\% & \multicolumn{1}{c}{OOT}  & $73.68_{\pm0.26}$  & $69.01_{\pm0.17}$  & -  & $\mathbf{77.02_{\pm0.56}}$  \\
\hline

\end{tabular}}
\vskip -0.2in
\end{table}

\subsubsection{\textbf{Against DICE.}}
To verify the defense generalization capability of LLM4RGNN, we also evaluate its effectiveness against another non-targeted attack, DICE~\cite{waniek2018hiding}. Notably, DICE is not involved in the construction process of the instruction dataset. Considering that DICE is not as effective as Mettack, we set higher perturbation rates of 10\%, 20\% and 40\%. The results are reported in Table~\ref{tab: dice_gnn} and Table~\ref{tab: dice_framework}. Similar to the results under Mettack, LLM4RGNN consistently improves the robustness across various GNNs and is superior to other robust GNN frameworks. For GCN, GAT, RGCN and Simp-GCN, LLM4RGNN on average brings 8.2\%, 8.8\%, 8.1\% and 6.5\% relative improvements in accuracy on five datasets. Remarkably, even in some cases where the perturbation ratio increases to 40\%, the accuracy of GNNs is better than that on the clean graph.

\subsubsection{\textbf{Against Nettack.}}
\begin{figure*}[htbp]
    \centering
    \begin{minipage}{0.48\linewidth}
        \centering
        \includegraphics[width=\linewidth]{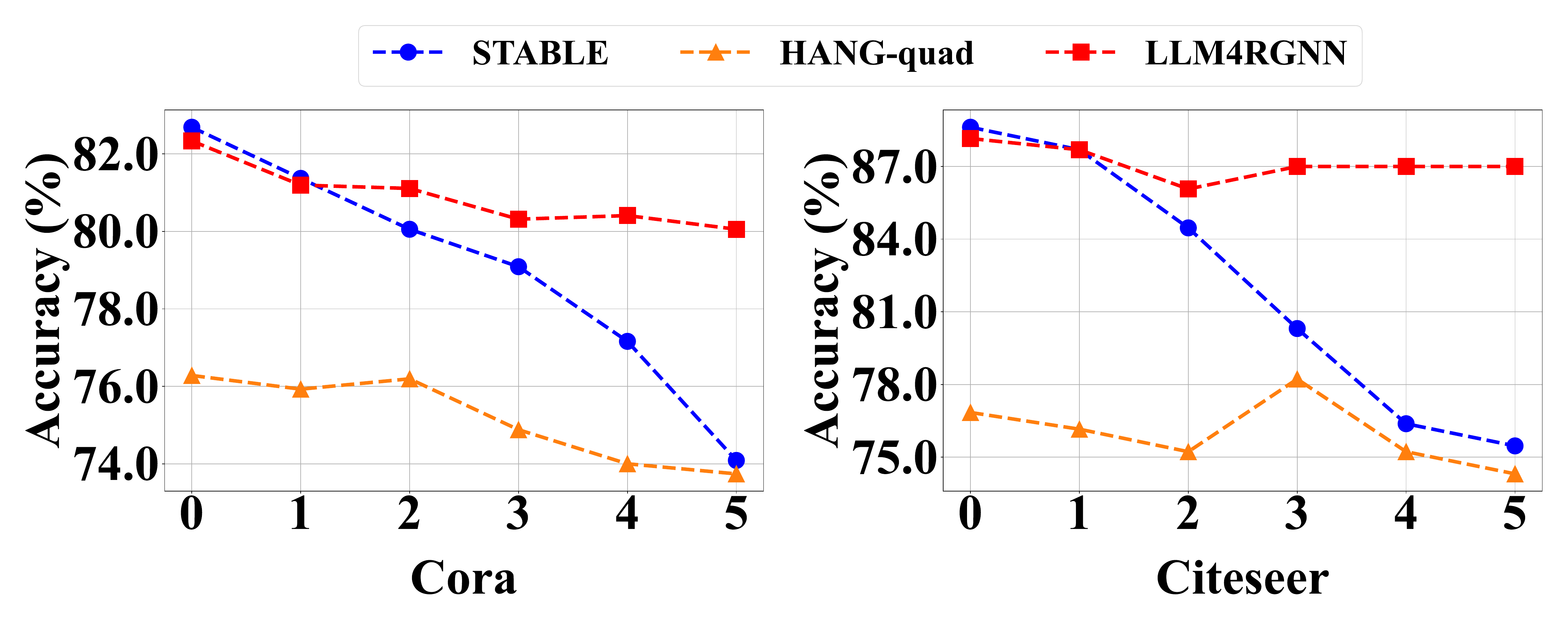}
        \vspace{-25pt}
        \caption{Node classification accuracy under targeted attack (Nettack) with different perturbations.} \label{fig: nettack_framework}
        \vskip -0.2in
    \end{minipage}
    \hfill
    \begin{minipage}{0.48\linewidth}
        \centering
       \includegraphics[width=\linewidth]{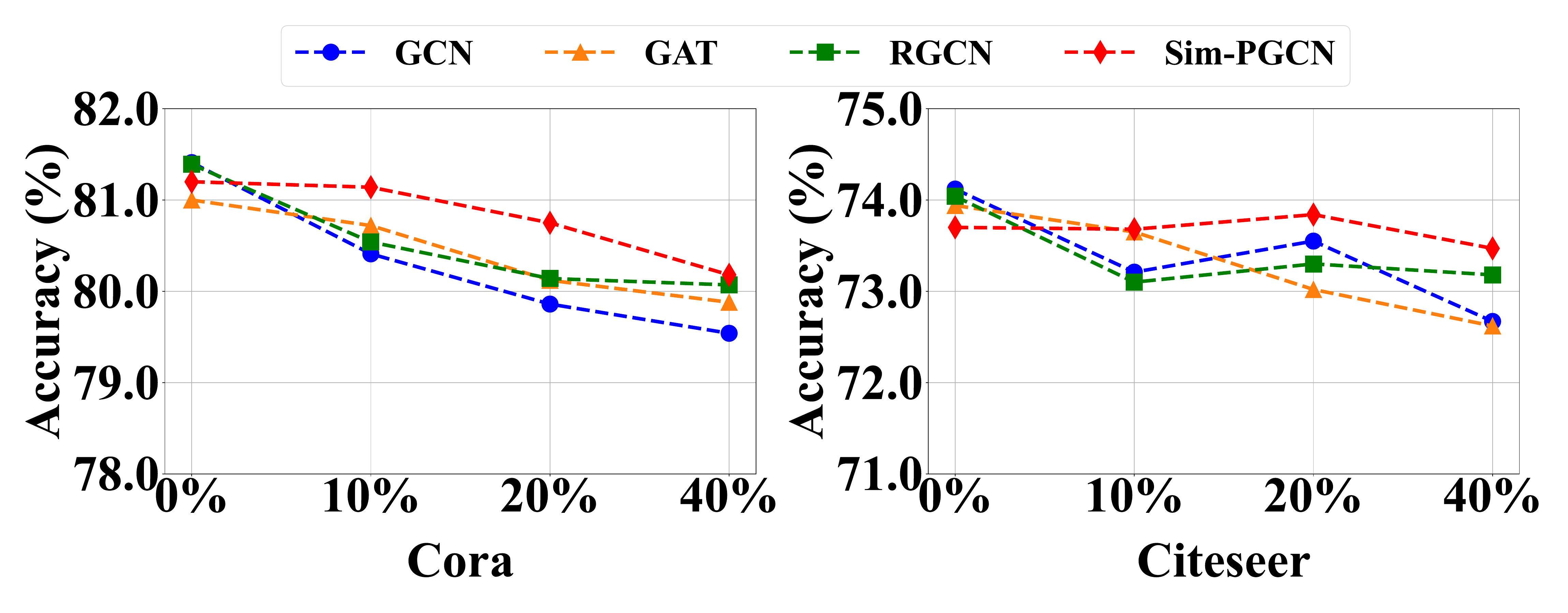}
        \vspace{-25pt}
        \caption{Node classification accuracy of LLM4RGNN under Mettack-20\% with different text perturbations.} \label{fig: text_attack}
        \vskip -0.2in
    \end{minipage}
\end{figure*}

Unlike non-targeted attacks, targeted attacks focus on fooling GNNs into misclassifying a particular node $v$. Here, we employ the SOTA targeted attack, Nettack~\cite{zugner2018adversarial}. Following previous work~\cite{zhu2019robust}, nodes with a degree greater than 10 are selected as the target nodes, and the number of perturbations applied to the targeted node from 0 to 5 with a step of 1. As shown in Figure~\ref{fig: nettack_framework} and Table~\ref{tab: nettack_gnn}, results indicate that LLM4RGNN not only consistently improves the robustness of various GNNs, but also surpasses existing robust frameworks, exhibiting exceptional resistance to Nettack.
\subsubsection{\textbf{Against Adaptive Attack.}}\label{exp: adaptive_attack}
Consider the worst-case scenario, where attackers have access to the well-tuned Mistral-7B. Thus, attackers can adaptively generate malicious edges that Mistral-7B rating >= 4, to avoid being removed. Based on Pubmed with more edges, we control the perturbation ratios at 5\% and 10\%. As reported in Tabel~\ref{tab: adaptive_attack}, the results indicate that LLM4RGNN effectively defends against adaptive attack and consistently improves the robustness of various GNNs. Although malicious edges \( r_e \in \{4, 5, 6\}\) failed to be removed, they are less aggressive than those \( r_e \in \{1, 2, 3\}\), 
and the important edges can further mitigate their impact.
\begin{table*}[t]
\caption{Node classification accuracy ($\%\pm\sigma$) under targeted attack (Nettack). Bolded results indicate improved performance.}\label{tab: nettack_gnn}
\vskip -0.15in
\centering
\resizebox{\linewidth}{!}{
\begin{tabular}{p{0pt}c|c>{\columncolor{myshadow}}c|c>{\columncolor{myshadow}}c|c>{\columncolor{myshadow}}c|c>{\columncolor{myshadow}}c} 
\hline
& \multicolumn{1}{c|}{\multirow{2}{*}{\makecell{Dataset \\ Ptb Num}}} & \multicolumn{2}{c|}{GCN} & \multicolumn{2}{c|}{GAT} & \multicolumn{2}{c|}{RGCN} & \multicolumn{2}{c}{Sim-PGCN} \\ 
\cline{3-10}
&                          & \multicolumn{1}{c}{Vanilla} & \multicolumn{1}{c|}{LLM4RGNN} & \multicolumn{1}{c}{Vanilla} & \multicolumn{1}{c|}{LLM4RGNN} & \multicolumn{1}{c}{Vanilla} & \multicolumn{1}{c|}{LLM4RGNN} & \multicolumn{1}{c}{Vanilla} & \multicolumn{1}{c}{LLM4RGNN} \\ 
\hline

\multirow{6}{*}{\hspace{-7pt} \rotcell{\makebox[8pt][c]{Cora}}}     & 0 & $85.92_{\pm2.21}$ & $\mathbf{87.50_{\pm1.79}}$ & $85.66_{\pm1.90}$ & $\mathbf{87.11_{\pm1.64}}$ & $86.45_{\pm1.87}$ & $\mathbf{88.55_{\pm2.57}}$ & $86.97_{\pm1.09}$ & $86.71_{\pm2.39}$ \\
                                    & 1 & $79.87_{\pm2.50}$ & $\mathbf{85.79_{\pm1.75}}$ & $82.11_{\pm2.06}$ & $\mathbf{86.71_{\pm2.24}}$ & $82.50_{\pm1.87}$ & $\mathbf{87.37_{\pm2.64}}$ & $81.71_{\pm2.39}$ & $\mathbf{86.05_{\pm2.37}}$ \\
                                    & 2 & $76.32_{\pm1.77}$ & $\mathbf{85.66_{\pm2.08}}$ & $75.13_{\pm4.42}$ & $\mathbf{86.18_{\pm1.79}}$ & $77.37_{\pm1.29}$ & $\mathbf{87.37_{\pm1.58}}$ & $77.76_{\pm1.37}$ & $\mathbf{84.34_{\pm1.99}}$ \\
                                    & 3 & $71.45_{\pm2.43}$ & $\mathbf{84.47_{\pm1.15}}$ & $70.79_{\pm3.85}$ & $\mathbf{85.00_{\pm2.58}}$ & $72.11_{\pm1.64}$ & $\mathbf{86.32_{\pm2.44}}$ & $71.97_{\pm2.89}$ & $\mathbf{85.39_{\pm1.37}}$ \\
                                    & 4 & $64.47_{\pm3.38}$ & $\mathbf{84.61_{\pm2.04}}$ & $65.79_{\pm3.86}$ & $\mathbf{85.39_{\pm1.99}}$ & $67.24_{\pm1.90}$ & $\mathbf{86.84_{\pm1.95}}$ & $63.55_{\pm3.58}$ & $\mathbf{84.61_{\pm2.50}}$ \\
                                    & 5 & $55.13_{\pm4.01}$ & $\mathbf{84.08_{\pm1.61}}$ & $60.00_{\pm4.68}$ & $\mathbf{83.03_{\pm1.90}}$ & $60.26_{\pm2.48}$ & $\mathbf{85.13_{\pm1.56}}$ & $52.63_{\pm4.40}$ & $\mathbf{83.55_{\pm1.69}}$ \\ 
\hline
\multirow{6}{*}{\hspace{-7pt} \rotcell{\makebox[8pt][c]{Citeseer}}} & 0 & $86.31_{\pm1.28}$ & $\mathbf{86.77_{\pm1.02}}$ & $86.92_{\pm0.77}$ & $\mathbf{87.38_{\pm1.02}}$ & $85.23_{\pm1.41}$ & $\mathbf{86.92_{\pm1.72}}$ & $80.92_{\pm2.50}$ & $\mathbf{84.62_{\pm2.84}}$ \\
                                    & 1 & $84.62_{\pm2.57}$ & $\mathbf{86.46_{\pm0.62}}$ & $85.54_{\pm1.23}$ & $ \mathbf{86.92_{\pm1.03}}$ & $82.15_{\pm1.85}$ & $\mathbf{86.31_{\pm1.45}}$ & $80.00_{\pm3.30}$ & $\mathbf{84.31_{\pm2.36}}$ \\
                                    & 2 & $81.69_{\pm2.62}$ & $\mathbf{85.38_{\pm1.03}}$ & $83.38_{\pm1.79}$ & $\mathbf{86.38_{\pm1.61}}$ & $74.46_{\pm4.25}$ & $\mathbf{85.85_{\pm1.51}}$ & $79.54_{\pm4.30}$ & $\mathbf{84.62_{\pm1.54}}$ \\
                                    & 3 & $79.08_{\pm3.24}$ & $\mathbf{86.00_{\pm0.63}}$ & $81.38_{\pm3.11}$ & $\mathbf{85.38_{\pm1.42}}$ & $72.00_{\pm3.94}$ & $\mathbf{85.54_{\pm1.41}}$ & $76.62_{\pm5.99}$ & $\mathbf{83.85_{\pm2.31}}$ \\
                                    & 4 & $76.77_{\pm5.26}$ & $\mathbf{86.00_{\pm0.83}}$ & $77.08_{\pm4.54}$ & $\mathbf{85.69_{\pm1.55}}$ & $69.08_{\pm2.70}$ & $\mathbf{85.38_{\pm1.58}}$ & $73.23_{\pm7.51}$ & $\mathbf{84.00_{\pm1.41}}$ \\
                                    & 5 & $74.31_{\pm5.29}$ & $\mathbf{86.00_{\pm1.08}}$ & $73.69_{\pm8.44}$ & $\mathbf{85.54_{\pm1.02}}$ & $64.31_{\pm3.94}$ & $\mathbf{84.46_{\pm1.45}}$ & $68.31_{\pm10.62}$ & $\mathbf{82.46_{\pm2.20}}$ \\
\hline
\end{tabular}}
\vskip -0.15in
\end{table*}

\begin{figure*}[htbp]
     \vskip -0.175in
     \centering
     \includegraphics[width=1\linewidth]{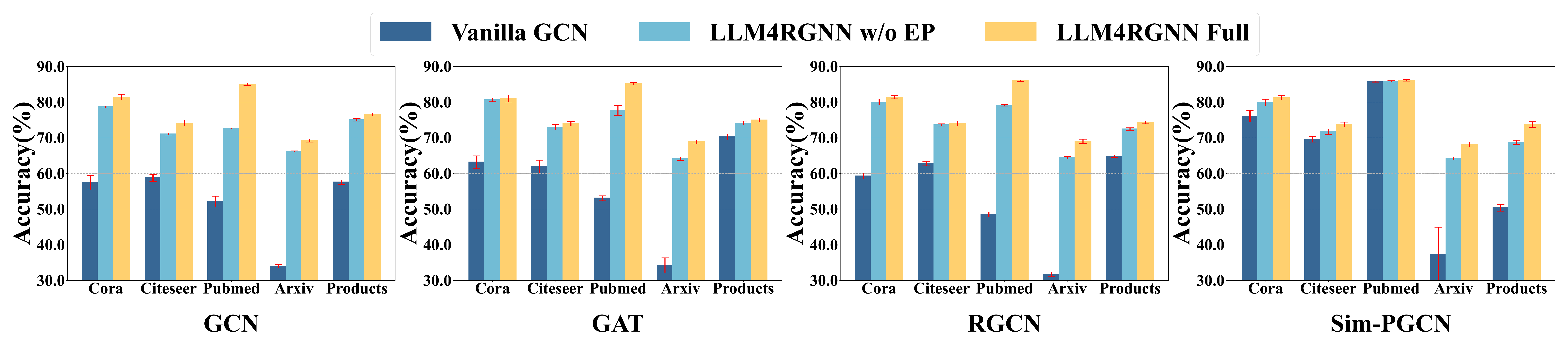}
     \vspace{-25pt}
     \caption{Performance comparison in different settings against Mettack with a 20\% perturbation rate.}
     \label{fig: ablation}
     \vskip -0.175in
\end{figure*}
\begin{table}[htp]
\caption{Node classification accuracy of Pubmed under adaptive attack. Bolded results indicate improved performance.} \label{tab: adaptive_attack}
\Huge
\vskip -0.15in
\setlength{\extrarowheight}{1pt}
\resizebox{\linewidth}{!}{
\centering
\begin{tabular}{c|cc|cc|cc}
\hline
\multirow{2}{*}{\makecell{Ptb \\ Rate}} & \multicolumn{2}{c|}{GCN}  & \multicolumn{2}{c|}{RGCN} & \multicolumn{2}{c}{SimP-GCN} \\ \cline{2-7} 
                          & Vanilla     & LLM4RGNN   & Vanilla     & LLM4RGNN    & Vanilla       & LLM4RGNN     \\ \hline
0\%                        & $85.62_{\pm0.10}$  & \cellcolor{gray!25}$\mathbf{86.21_{\pm0.13}}$ & $85.84_{\pm0.10}$  & \cellcolor{gray!25}$\mathbf{86.35_{\pm0.10}}$  & $87.26_{\pm0.08}$    & \cellcolor{gray!25}$\mathbf{87.53_{\pm0.14}}$   \\
5\%                        & $84.23_{\pm0.12}$  & \cellcolor{gray!25}$\mathbf{84.70_{\pm0.28}}$ & $84.49_{\pm0.10}$  & \cellcolor{gray!25}$\mathbf{85.43_{\pm0.20}}$  & $86.21_{\pm0.10}$    & \cellcolor{gray!25}$\mathbf{86.36_{\pm0.25}}$   \\
10\%                       & $81.51_{\pm0.20}$  & \cellcolor{gray!25}$\mathbf{84.82_{\pm0.26}}$ & $82.69_{\pm0.21}$  & \cellcolor{gray!25}$\mathbf{85.58_{\pm0.20}}$  & $85.86_{\pm0.16}$    & \cellcolor{gray!25}$\mathbf{86.11_{\pm0.16}}$   \\ \hline
\end{tabular}
}
\vskip -0.2in
\end{table}
\subsection{Model Analysis}
\subsubsection{\textbf{Ablation Study.}}
To assess the impact of key components in LLM4RGNN, we conduct ablation studies under Mettack with a 20\% perturbation rate, where GCN is selected as training GNN. The results are depicted in Figure~\ref{fig: ablation}, where "Vanilla" is without any modifications to the attacked topology. The "w/o EP" only removes malicious edges by the local LLM, while "Full" includes both removing malicious edges and adding important edges. Across all settings, LLM4RGNN Full consistently outperforms other settings. Using the local LLM to remove most malicious edges can significantly reduce the impact of adversarial perturbation, improving the accuracy of GNNs. By employing the edge predictor to add important neighbors for each node, additional information gain is provided to the center nodes, further improving the accuracy of GNNs.
\subsubsection{\textbf{Impact of Text Quality.}}\label{exp: text_attack}
Considering that LLM4RGNN relies on textual information for reasoning, we analyze the impact of text quality on the model's effectiveness. Specifically, against the worst-case scenario of 20\% Mettack, we further add random text replacement perturbations to the Cora and Citeseer to reduce the text quality of nodes. As shown in Figure~\ref{fig: text_attack}, results show that LLM4RGNN's performance varies by less than two points under 10-40\% perturbations, and surpasses existing robust GNN frameworks. A possible explanation is that LLMs have strong robustness to text perturbations~\cite{chang2024survey}, and LLM4RGNN fully inherits it.

\begin{table*}[t]
\caption{Performance comparison with different LLMs against Mettack with a 20\% perturbation rate.}\label{tab: diff_llms}
\vskip -0.15in
\centering
\resizebox{\linewidth}{!}{
\begin{tabular}{cc|ccc|ccc} 
\hline
\multicolumn{2}{c|}{\multirow{2}{*}{LLM}} & \multicolumn{3}{c|}{Cora}                                                                                             & \multicolumn{3}{c}{Citeseer}            
\\
\cline{3-8}

\multicolumn{2}{c|}{}                                      & \multicolumn{1}{c}{AdvEdge (↓)} & \multicolumn{1}{c}{ACC (↑) w/o EP} & \multicolumn{1}{c|}{ACC (↑) Full} & \multicolumn{1}{c}{AdvEdge (↓)} & \multicolumn{1}{c}{ACC (↑) w/o EP} & \multicolumn{1}{c}{ACC (↑) Full}  \\ 

\hline
\multirow{2}{*}{\shortstack{Close\\source}} & GPT-3.5                     & $186(3.52\%)$                & $73.56_{\pm 0.49}$                                 & $75.36_{\pm 1.73}$                                 & $231(5.47\%)$                 & $66.69_{\pm 1.04}$                                & $68.92_{\pm 1.63}$                                \\
                              & GPT-4                     & $\mathbf{90(1.71\%)}$                 & $\mathbf{79.83_{\pm 0.31}}$                                 & $\mathbf{81.76_{\pm 0.87}}$                                 & $\mathbf{73(1.73\%)}$                  & $\mathbf{71.04_{\pm 0.57}}$                                & $\mathbf{72.81_{\pm 0.60}}$                                \\
\hline
\multirow{4}{*}{\shortstack{Open\\source}}  & Llama2-7B                 & $156(2.96\%)$                 & $78.76_{\pm 0.42}$                                 & $80.99_{\pm 0.78}$                                 & $203(4.80\%)$                 & $69.83_{\pm 0.78}$                                & $72.86_{\pm 0.93}$                                \\
                              & Llama2-13B                & $132(2.50\%)$                 & $79.00_{\pm 0.30}$                                 & $81.35_{\pm 1.15}$                                 & $93(2.20\%)$                  & $70.81_{\pm 0.73}$                                & $72.84_{\pm 0.85}$                                \\
                              & Llama3-8B                 & $107(2.03\%)$                 & $\mathbf{79.22_{\pm 0.34}}$                                 & $\mathbf{81.71_{\pm 0.51}}$                                 & $\mathbf{91(2.15\%)}$                  & $70.85_{\pm 0.64}$                                & $74.06_{\pm 0.78}$                                \\
                              & Mistral-7B                 & $\mathbf{102(1.93\%)}$                 & $78.68_{\pm 0.28}$                                 & $81.41_{\pm 0.77}$                                 & $92(2.18\%)$                  & $\mathbf{71.07_{\pm 0.35}}$                                & $\mathbf{74.12_{\pm 0.85}}$                                \\
\hline
\end{tabular}}
\vskip -0.1in
\end{table*}

\begin{figure*}[tb]
\vskip -0.085in
\centering
\subfigure[\textbf{Cora-Mettack-5\%.}]
{
    \begin{minipage}[b]{.22\linewidth}
        \centering
        \includegraphics[width=\textwidth, trim={0 0 0 2.5cm}, clip]{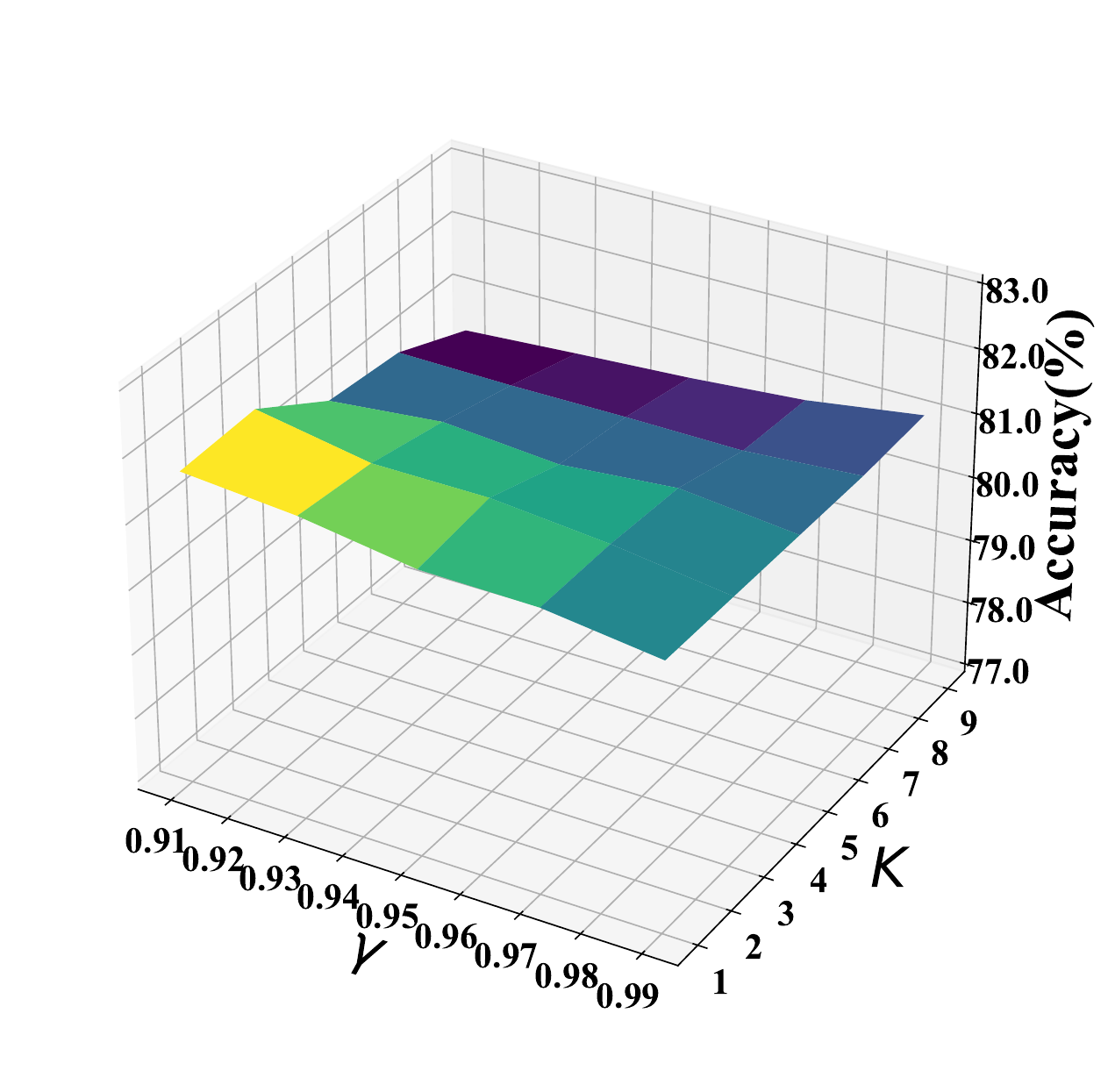}
        \vspace{-22pt}
    \end{minipage}
}
\subfigure[\textbf{Cora-Mettack-20\%.}]
{
 	\begin{minipage}[b]{.22\linewidth}
        \centering
        \includegraphics[width=\textwidth, trim={0 0 0 2.5cm}, clip]{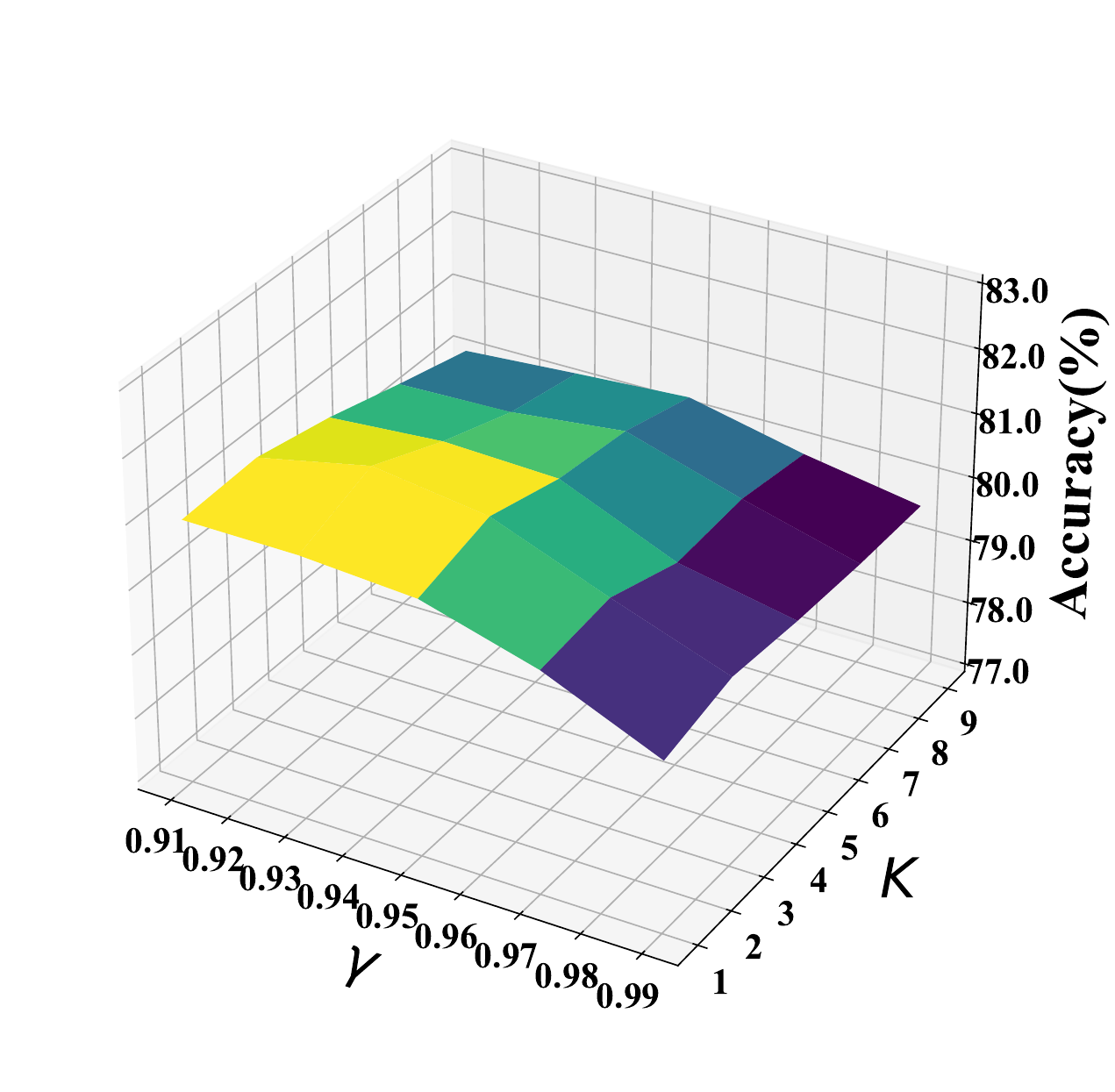}
        \vspace{-22pt}
    \end{minipage}
}
\subfigure[\textbf{Citeseer-Mettack-5\%.}]
{
    \begin{minipage}[b]{.22\linewidth}
        \centering
        \includegraphics[width=\textwidth, trim={0 0 0 2.5cm}, clip]{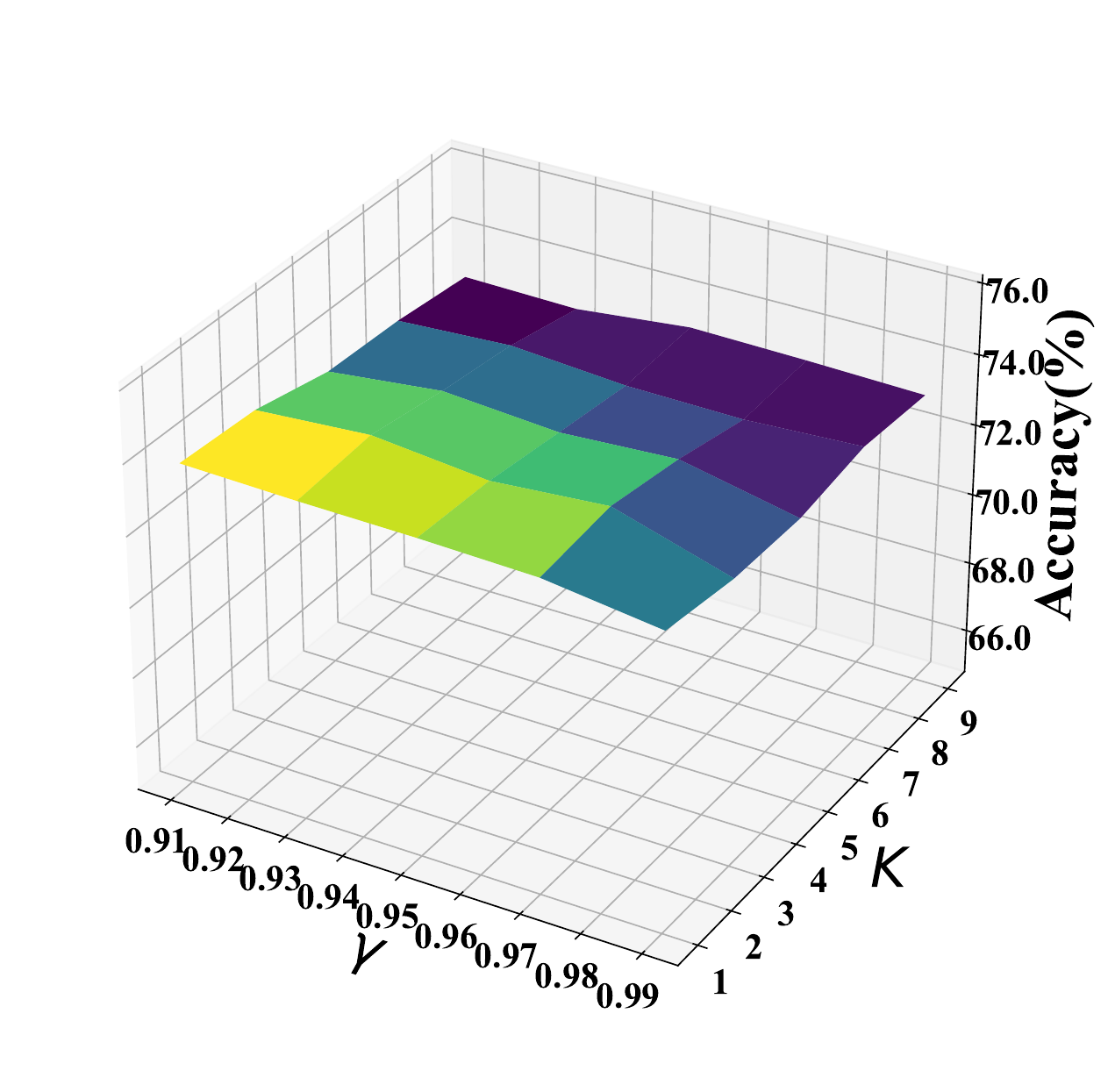}
        \vspace{-22pt}
    \end{minipage}
}
\subfigure[\textbf{Citeseer-Mettack-20\%.}]
{
 	\begin{minipage}[b]{.22\linewidth}
        \centering
        \includegraphics[width=\textwidth, trim={0 0 0 2.5cm}, clip]{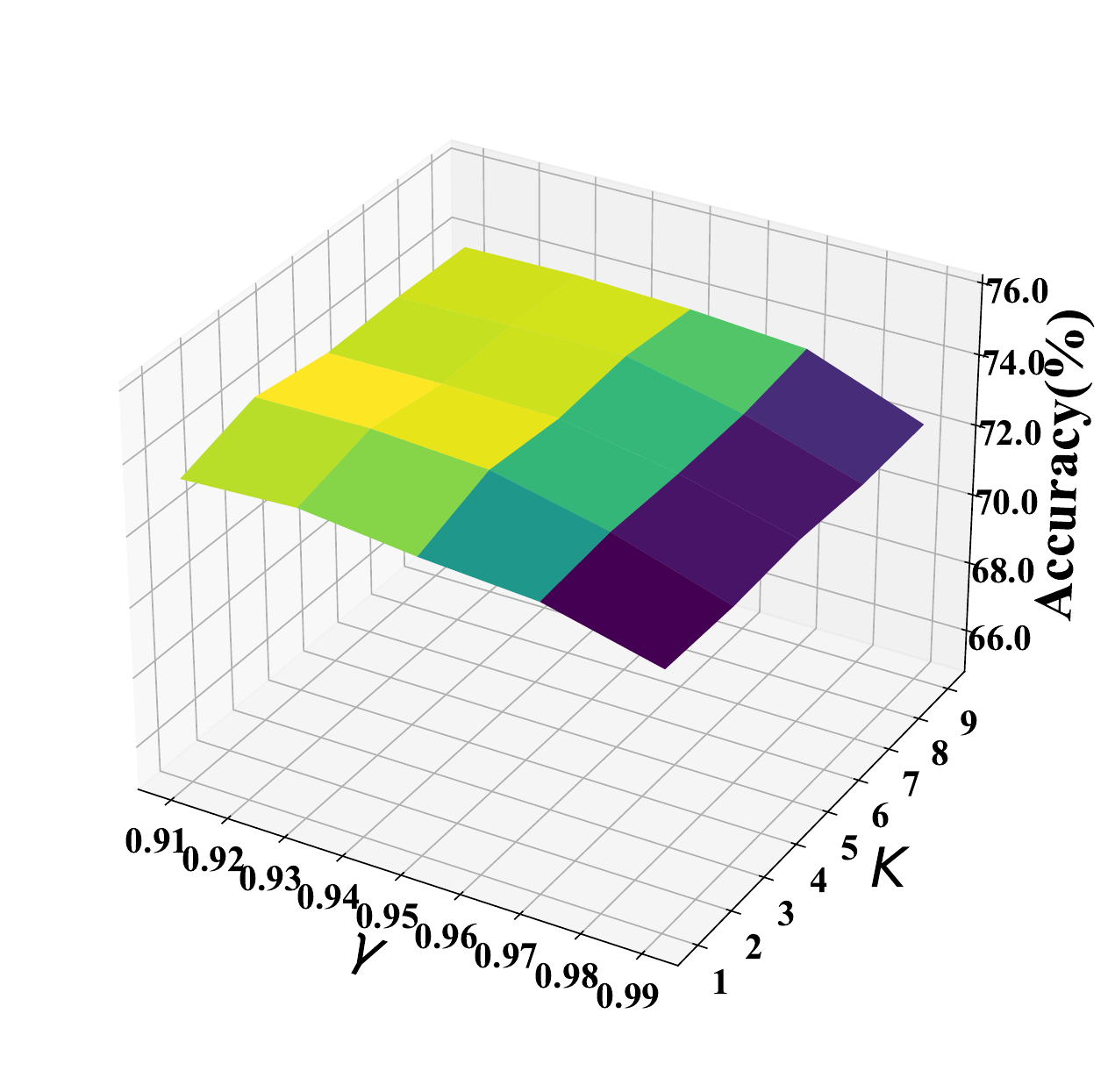}
        \vspace{-22pt}
    \end{minipage}
}
    \vskip -0.2in
    \caption{Analysis of the hyper-parameter $\gamma$ and $K$ against Mettack.}
    \label{fig: hyper_gamma_k_cora_citeseer_5_20}
\end{figure*}

\begin{figure*}[htbp]
    \centering
    \begin{minipage}{0.48\linewidth}
        \centering
        \includegraphics[width=\linewidth]{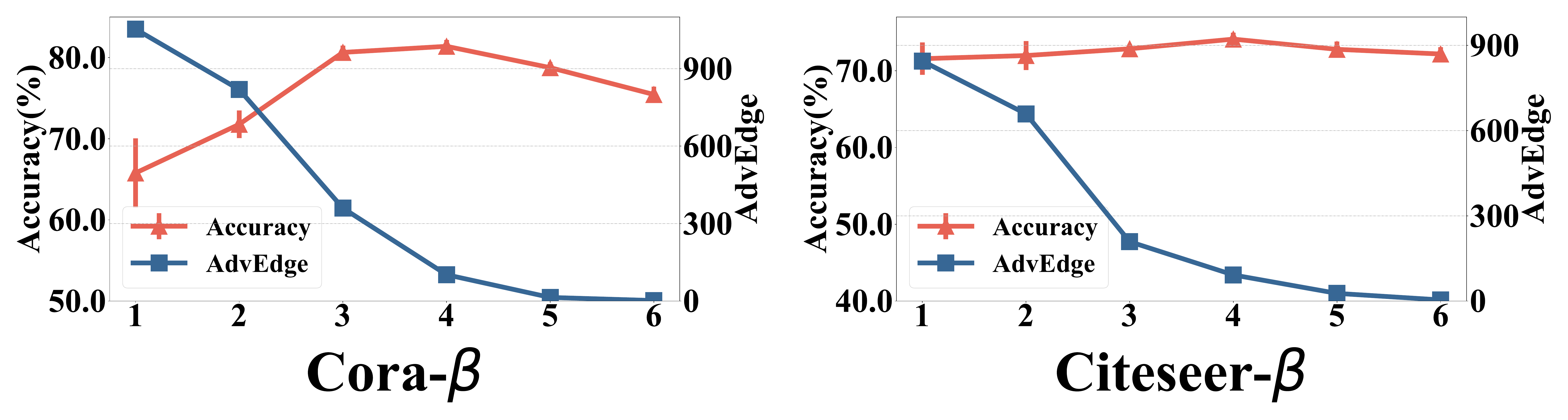}
        \vspace{-21pt}
        \caption{Analysis of hyper-parameter $\beta$ against Mettack-20\%.}
        \label{fig: hyper_beta}
        \vspace{-15pt}
    \end{minipage}
    \hfill
    \begin{minipage}{0.48\linewidth}
        \centering
       \vspace{-11pt} 
       \includegraphics[width=\linewidth]{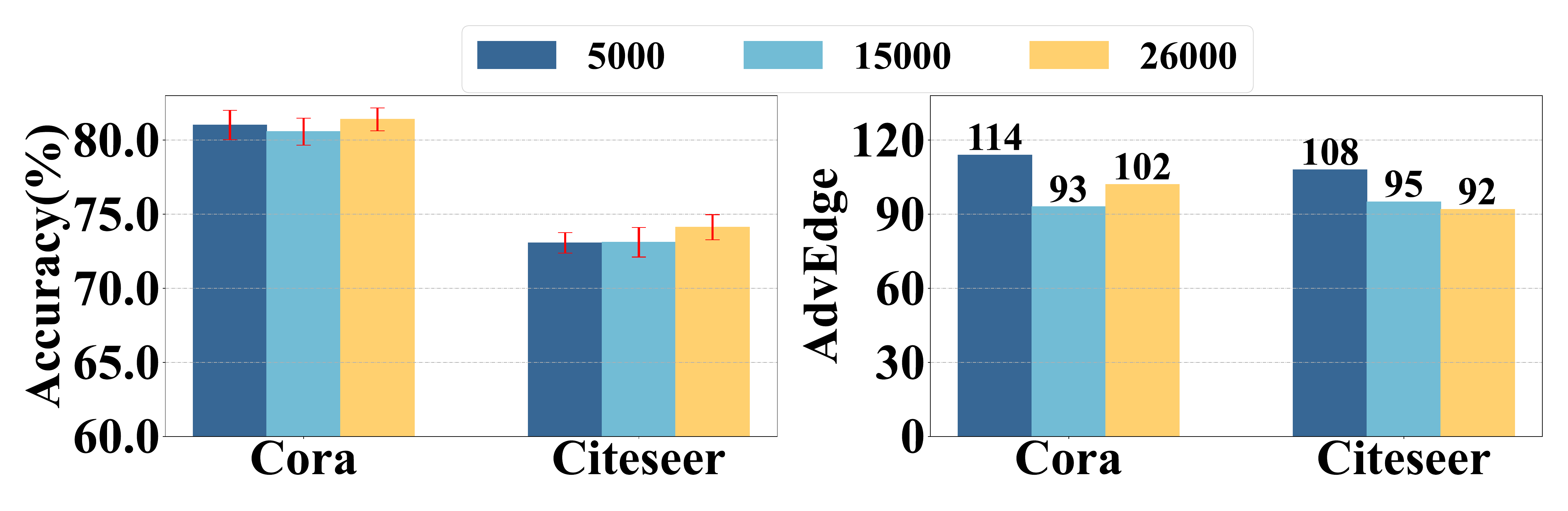}
        \vspace{-25pt}
        \caption{Analysis of instances number against Mettack-20\%.}
        \vspace{-15pt}
        \label{fig: hyper_instance}
    \end{minipage}
\end{figure*}

\subsubsection{\textbf{Comparison with Different LLMs.}}\label{exp: diff_llm}
To evaluate the generalizability of LLM4RGNN across different LLMs, we choose four popular open-source LLMs, including Llama2-7B, Llama2-13B, Llama3-8B and Mistral-7B, as the starting checkpoints of LLM. 
We also introduce a direct comparison using the closed-source GPT-3.5 and GPT-4. Additionally, the metric AdvEdge (↓) is introduced to measure the number and proportion of malicious edges remaining after the LLM performs the filtering operation. 
We report the results of GCN on Cora and Citeseer under Mettack with a perturbation ratio of 20\% (generating 1053 malicious edges for Cora and 845 for Citeseer). 
As reported in Table~\ref{tab: diff_llms}, we observe: (1) The well-tuned local LLMs are significantly superior to GPT-3.5 in identifying malicious edges, and the trained LM-based edge predictor consistently improves accuracy, indicating that the inference capability of GPT-4 is effectively distilled into different LLMs and edge predictors.
(2) A stronger open-source LLM yields better overall performance. Among them, the performance of fine-tuned Mistral-7B and Llama3-8B is comparable to that of GPT-4. We also provide a comprehensive comparison between them in Appendix~\ref{app: llama_mistral_compare}.

\subsubsection{\textbf{Hyper-parameter Analysis.}}\label{exp: hyper-parameter}
First, we present the accuracy of LLM4RGNN under different combinations of the probability threshold $\lambda$ and the maximum number of important edges $K$ in Figure~\ref{fig: hyper_gamma_k_cora_citeseer_5_20}. The results indicate that the accuracy of LLM4RGNN varies minimally across different hyper-parameter settings, demonstrating its insensitivity to the hyper-parameters $\lambda$ and $K$. More results refer to Appendix~\ref{sec: exp-hyper-details}. Besides, we report the accuracy and AdvEdge of LLM4RGNN under different $\beta$, the purification threshold of whether an edge is preserved or not. As shown in Figure~\ref{fig: hyper_beta}, when the $\beta$ is set to 4, most malicious edges are identified and achieve optimal performance. This is because a low $\beta$ fails to identify malicious edges effectively, while a high $\beta$ removes more malicious edges but could remove some useful edges. Lastly, we set the number of instances for tuning the local LLM to 5000, 15000, and 26000, respectively, to analyze the model’s effectiveness with different instance counts. As shown in Figure~\ref{fig: hyper_instance}, with only 5000 instances, the tuned local LLM can effectively identify malicious edges, surpassing the existing SOTA method, which indicates that the excellent robustness of LLM4RGNN can be achieved on a lower budget, approximately \$46.7. Detailed tokens cost refer to Appendix~\ref{exp: resource_cost}.

\begin{table}[htp]
    \centering
    \vskip -0.1in
    \caption{Average inference times (s) for different datasets.}
    \vskip -0.15in
    \label{tab:inference_times}
    \resizebox{\linewidth}{!}{
    \begin{tabular}{c|ccccc}
        \hline
        Component & Cora & Citeseer & Products & Pubmed & Arxiv \\
        \hline
        Local LLM & 0.54 & 0.57 & 0.63 & 0.64 & 1.12 \\
        Edge Predictor & 0.05 & 0.06 & 0.24 & 0.31 & 0.25 \\
        \hline
    \end{tabular}
    }
    \vskip -0.15in
\end{table}
\begin{table*}[tp]
\caption{Statistics of edges purified by LLM4RGNN. We also report the percentage of malicious edges removed.}\label{tab: restore_edge_statistic}
\vskip -0.15in
\setlength{\extrarowheight}{-0.5pt}
\resizebox{\linewidth}{!}{
\begin{tabular}{c|c|c|c|c|c|c|c}
\hline
\multicolumn{1}{c|}{\multirow{2}{*}{Dataset}} & \multirow{2}{*}{Type} & \multicolumn{3}{c|}{Mettack}            & \multicolumn{3}{c}{DICE}                \\ \cline{3-8} 
\multicolumn{1}{c|}{}                         &                       & 5\%          & 10\%         & 20\%         & 10\%         & 20\%         & 40\%         \\ \hline
\multirow{2}{*}{Cora}                          & Remove                & $234(89.3\%)$  & $466(88.6\%)$  & $951(90.3\%)$  & $216(87.8\%)$  & $429(86.1\%)$  & $932(88.9\%)$  \\
                                               & Add                   & $9,362$        & $6,440$        & $9,878$        & $6,307$        & $5,928$        & $8,495$        \\ \hline
\multirow{2}{*}{Citeseer}                      & Remove                & $181(85.8\%)$  & $367(87.0\%)$  & $753(89.1\%)$  & $184(91.1\%)$  & $365(93.6\%)$  & $753(91.7\%)$  \\
                                               & Add                   & $12,464$       & $14,332$       & $7,215$        & $10,569$       & $12,434$       & $11,372$       \\ \hline
\multirow{2}{*}{Pubmed}                        & Remove                & $1,401(63.2\%)$ & $2,708(61.1\%)$ & $4,972(56.9\%)$ & $1,188(53.7\%)$ & $2,281(51.3\%)$ & $4,749(53.6\%)$ \\
                                               & Add                   & $56,505$       & $100,122$      & $97,057$       & $57,595$       & $96,617$       & $91,525$       \\ \hline
\multirow{2}{*}{Arxiv}                         & Remove                & $799(95.5\%)$  & $1,600(95.6\%)$ & $3,204(96.3\%$) & $786(96.7\%)$  & $1,629(96.3\%)$ & $3,227(95.9\%)$ \\
                                               & Add                   & $49,741$       & $52,253$       & $52,621$       & $53,190$       & $47,046$       & $46,807$       \\ \hline
\multirow{2}{*}{Products}                      & Remove                & $483(65.2\%)$  & $1,001(67.5\%)$ & $2,059(69.5\%)$ & $679(94.8\%)$  & $1,400(94.2\%)$ & $2,799(94.4\%)$ \\
                                               & Add                   & $25,413$       & $22,731$       & $46,827$       & $24,177$       & $44,041$       & $96,945$       \\ \hline
\end{tabular}
}
\vskip -0.1in
\end{table*}
\subsubsection{\textbf{Efficiency Analysis.}}\label{exp: efficiency}
In LLM4RGNN, using 26,518 samples to fine-tune the local LLM is a one-time process, controlled within 9 hours. The edge predictor is only a lightweight MLP, with training time on each dataset controlled within 3 minutes. The complexity of LLM inferring edge relationships is $O(\mathcal{|E|})$, and the edge predictor is $O({|\mathcal{V}|}^2)$, and their inference processes are parallelizable. We provide the average time for LLM to infer one edge and for the lightweight edge predictor to infer one node in Table~\ref{tab:inference_times}. Overall, the average inference time for the local LLM and the edge predictor is 0.7s and 0.2s, respectively, which is acceptable. Each experiment's total inference time is controlled within 90 minutes. Here, we further discuss how to extend LLM4RGNN to large-scale graphs. For the local LLM with complexity $O(\mathcal{E}|)$, we introduce the parallel inference framework vLLM~\cite{kwon2023efficient} and cache the edges inferred by the LLM. For the edge predictor with complexity $O({|\mathcal{V}|}^2)$, we infer only the top $K$ most similar nodes for each node (where $K$ << $|\mathcal{V}|$), to reduce the complexity to $O({K|\mathcal{V}|})$. As reported in Table~\ref{tab: full_arxiv}, results on complete OGBN-Arxiv (169,343 nodes and 1,166,243 edges) indicate the effectiveness of LLM4RGNN on the large-scale graph, with an average inference time of 7 hours (where $K$ is set to 2000).
\begin{table}[tp]
\caption{Performance of complete OGBN-Arxiv under DICE. Bolded results indicate improved performance.} \label{tab: full_arxiv}
\Huge
\vskip -0.15in
\resizebox{\linewidth}{!}{
\begin{tabular}{c|cc|cc|cc}
\hline
\multirow{2}{*}{\makecell{Ptb \\ Rate}} & \multicolumn{2}{c|}{GCN} & \multicolumn{2}{c|}{GAT} & \multicolumn{2}{c}{SimP-GCN} \\ \cline{2-7} 
                          & Vanilla     & LLM4RGNN   & Vanilla     & LLM4RGNN   & Vanilla       & LLM4RGNN     \\ \hline
0\%                        & $68.33_{\pm0.04}$  & \cellcolor{gray!25}$\mathbf{68.82_{\pm0.14}}$ & $64.99_{\pm0.10}$  & \cellcolor{gray!25}$\mathbf{66.48_{\pm0.16}}$ & $64.50_{\pm3.51}$    & \cellcolor{gray!25}$\mathbf{65.54_{\pm3.94}}$   \\
10\%                       & $65.22_{\pm0.05}$  & \cellcolor{gray!25}$\mathbf{69.26_{\pm0.04}}$ & $61.14_{\pm0.08}$  & \cellcolor{gray!25}$\mathbf{67.85_{\pm0.11}}$ & $61.41_{\pm4.76}$    & \cellcolor{gray!25}$\mathbf{67.25_{\pm2.92}}$   \\
20\%                       & $62.13_{\pm0.09}$  & \cellcolor{gray!25}$\mathbf{68.80_{\pm0.07}}$ & $57.77_{\pm0.13}$  & \cellcolor{gray!25}$\mathbf{67.46_{\pm0.08}}$ & $59.39_{\pm4.68}$    & \cellcolor{gray!25}$\mathbf{66.74_{\pm3.16}}$   \\
40\%                       & $56.75_{\pm0.05}$  & \cellcolor{gray!25}$\mathbf{68.40_{\pm0.05}}$ & $52.04_{\pm0.11}$  & \cellcolor{gray!25}$\mathbf{66.78_{\pm0.09}}$ & $55.65_{\pm2.09}$    & \cellcolor{gray!25}$\mathbf{66.26_{\pm2.90}}$   \\ \hline
\end{tabular}
}
\vskip -0.2in
\end{table}

\subsubsection{\textbf{Purifying Edge Statistic.}}\label{exp: restore_edge_statistic}
To analyze how LLM4RGNN defends various attacks, we reported the number of edges added and removed, and the percentage of malicious edges removed in Table~\ref{tab: restore_edge_statistic}. We find that LLM4RGNN removes an average of 82.4\% of malicious edges across five datasets, and mitigates the impact of remaining malicious edges by adding massive important edges. An interesting observation is that the LLM fine-tuned on TAPE-Arxiv performs best at identifying malicious edges on OGBN-Arxiv, which indicates that using domain-specific LLMs can further enhance LLM4RGNN.

\section{Related Work}\label{append: related-work}
\subsection{Adversarial Attack and Defense on Graph}
Extensive studies~\cite{li2022revisiting, waniek2018hiding, mujkanovic2022defenses, jin2021adversarial, zhang2023minimum, zhang2022robust} have shown that attackers can catastrophically degrade the performance of GNNs by maliciously perturbing the graph structure. For example, Nettack~\cite{zugner2018adversarial} first to study the adversarial attacks on graph data, which preserves degree distribution and imposes constraints on feature co-occurrence to generate small deliberate perturbations. Subsequently, Mettack~\cite{zugner_adversarial_2019} utilizes meta-learning while Minmax and PGD~\cite{xu2019topology} utilize projected gradient descent, to solve the bilevel problem underlying poisoning attacks. Threatened by adversarial attacks, many methods~\cite{gosch2023revisiting, huang2023robust, li2023boosting, gosch2024adversarial} have been proposed to defend against adversarial attacks. They are mainly categorized into model-centric and data-centric. The methods of model-centric improve the robustness through model enhancement, either by robust training schemes (e.g., adversarial training~\cite{li2023boosting, gosch2024adversarial}) or designing new model architectures (e.g., RGCN~\cite{zhu2019robust}, HANG~\cite{zhao2024adversarial}, Mid-GCN\cite{huang2023robust}).
The methods of data-centric typically focus on flexible data processing. By treating the attacked topology as noisy, defenders primarily purify graph structures by calculating various similarities between node embeddings~\cite{wu2019adversarial, jin2020graph, zhang2020gnnguard, li2022reliable, entezari2020all}. For instance, ProGNN~\cite{jin2020graph} jointly trains GNN and learns a clean adjacency matrix with graph properties. STABLE~\cite{li2022reliable} learns effective representations by unsupervised pre-training to refine graph structures.

\subsection{LLMs for Graph}
Recently, LLMs have been widely employed in graph-related tasks, which even outperform traditional GNN-based methods. According to the role played by LLMs in graph-related tasks, some methods utilize LLMs as an enhancer~\cite{he2023harnessing, liu2023one, xu2024pefad}, to enhance the quality of node features. Some methods directly use LLMs as a predictor~\cite{wang2024can, chen2024exploring, zhao2024gimlet, zhang2024graphtranslator}, where the graph structure is described in natural language for input to LLMs for prediction. Additionally, some methods employ LLMs as an annotator~\cite{chen2023label}, generator~\cite{yu2023empower}, and controller~\cite{wang2023graph}. 
Notably, recent GraphEdit~\cite{guo2024graphedit} is a data augmentation framework, which aims to improve the performance ceiling on the original graph. In comparison, LLM4RGNN is a defense framework against topology attacks, which aims to improve the stability of various GNNs under attack. Although both focus on utilizing LLMs for graph structure learning, their motivations are distinct.
\section{Conclusion}
In this paper, we first explore the potential of LLMs on the graph adversarial robustness. Specifically, we propose a novel LLM-based robust graph structure inference framework, LLM4RGNN, which distills the inference capability of GPT-4 into a local LLM for identifying malicious edges and an LM-based edge predictor for finding missing important edges, to efficiently purify attacked graph structure, making GNNs more robust. Extensive experiments demonstrate that LLM4RGNN significantly improves various GNNs' robustness. Considering there are some graphs that lack textual information, a plan is to extend LLM4RGNN to graphs without text.
\begin{acks}
This work is supported by the National Key Research and Development Program of China (2023YFF0725103), the National Natural Science Foundation of China (U22B2038, 62322203, 62172052, 62192784), and 
the Young Elite Scientists Sponsorship Program (No.2023QNRC001) by CAST.
\end{acks}
\clearpage
\medskip
\bibliographystyle{plain}
\bibliography{reference}
\appendix
\clearpage

\section{Experiment Detail}
\subsection{Dataset}\label{sec: exp-dataset}
In this paper, we select the TAPE-Arxiv23~\cite{he2023harnessing} with up-to-date and rich texts to construct the instruction dataset, and use the five popular node classification datasets to evaluate model performance: Cora~\cite{mccallum2000automating}, Citeseer~\cite{giles1998citeseer}, Pubmed~\cite{sen2008collective}, OGBN-Arxiv~\cite{hu2020open} and OGBN-Products~\cite{hu2020open}. For each graph, following existing works~\cite{jin2020graph, li2022reliable}, we randomly split the nodes into 10\% for training, 10\% for validation, and 80\% for testing.
Notably, for OGBN-Arxiv (with 169,343 nodes and 1,166,243 edges) and OGBN-Products (with 2 million nodes and 61 million edges), in the main experiments, since we consider the strongest Mettack and Nettack, with high complexity, constructing their attack topologies is not feasible. Thus, we adopt a node sampling strategy~\cite{hamilton2017inductive} to obtain their subgraph. We give detailed statistics of each dataset in Table~\ref{tab:stats}. 

\begin{table}[t]
\centering
\caption{Dataset statistics.}
\vskip -0.15in
\Huge
\label{tab:stats}
\setlength{\extrarowheight}{1pt}
\resizebox{\linewidth}{!}{
\begin{tabular}{c|c|c|c|c|c}
\hline
\textbf{Dataset} & \textbf{\#Nodes} & \textbf{\#Edges} & \textbf{\#Classes} & \textbf{\#Features} & \textbf{Method}  \\
\hline
Cora & $2,708$ & $5,429$ & $7$ & $1,433$ & BoW  \\ \hline
Citeseer & $3,186$ & $4,225$ & $6$ & $3,113$ & BoW  \\ \hline
PubMed & $19,717$ & $44,338$ & $3$ & $500$ & TF-IDF  \\ \hline
OGBN-Arxiv (subset) & $14,167$ & $33,520$ & $40$ & $128$ & skip-gram \\ \hline
OGBN-Products (subset) & $12,394$ & $29,676$ & $47$ & $100$ & BoW \\ \hline
TAPE-Arxiv23 & $46,198$ & $78,548$ & $40$ & $300$ & word2vec  \\
\hline
\end{tabular}
}
\vskip -0.15in
\end{table}

\subsection{Baseline}\label{sec:exp-baseline}
\begin{itemize}[leftmargin=*]
    \item \textbf{GCN}: GCN is a popular graph convolutional network based on spectral theory.
    \item \textbf{GAT}: GAT is composed of multiple attention layers, which can learn different weights for different neighborhood nodes. It is often used as a baseline for defending against adversarial attacks.
    \item \textbf{RGCN}: RGCN models node representations as Gaussian distributions to mitigate the impact of adversarial attacks, and employs an attention mechanism to penalize nodes with high variance.
    \item \textbf{Simp-GCN}: SimpGCN employs a $k$NN graph to maintain the proximity of nodes with similar features in the representation space and uses self-learned regularization to preserve the remoteness of nodes with differing features.
    \item \textbf{ProGNN}: ProGNN adapts three regularizations of graphs, i.e., feature smoothness, low-rank, and sparsity, and learns a clean adjacency matrix to defend against adversarial attacks.
    \item \textbf{STABLE}: STABLE is a pre-training model, which specifically learns effective representations to refine graph structures.
    \item \textbf{HANG-quad}: HANG-quad incorporates conservative Hamiltonian flows with Lyapunov stability to various GNNs, to improve their robustness against adversarial attacks.
    \item \textbf{GraphEdit}: GraphEdit utilizes LLMs to identify noisy connections and uncover implicit relations among non-connected nodes in the original graph.
\end{itemize}
\subsection{Implementation Detail}\label{sec:exp-implementation}
For LLM4RGNN, we select Mistral-7B as our local LLM. Based on GPT-4, we construct approximately 26,518 instances for tuning LLMs and use the LoRA method to achieve parameter-efficient fine-tuning. To address the potential problem of label imbalance in training LM-based edge predictor, we select the 4,000 node pairs with the lowest cosine similarity to construct the candidate set.
We set the hyper-parameters of LLM4RGNN as follows: For local LLMs, when no purification occurs, the purification threshold \(\beta\) is selected from $\{1, 2\}$ to prevent deleting too many edges; otherwise, it is selected from $\{2, 3, 4\}$. For LM-based edge predictor, the threshold $\gamma$ is tuned from $\{0.91, 0.93, 0.95, 0.97, 0.99 \}$ and the number of edges $K$ is tuned from $\{1, 3, 5, 7, 9\}$. 

We use DeepRobust\cite{li2020deeprobust}, an adversarial attack repository, to implement all attacks as well as GCN, GAT, RGCN, and Sim-PGCN. We implement ProGNN, STABLE, and HANG-quad with the code provided by the authors. 
To facilitate fair comparisons, we tune all baselines' parameters using a grid search strategy. Unless otherwise specified, we adopt the default parameter setting in the author’s implementation. 
For GCN, the hidden size is 256 for OGBN-Arxiv and 16 for others. 
For GAT, the hidden size is 128 for OGBN-Arxiv and 8 for others.
For RGCN and Sim-PGCN, the hidden size is 256 for OGBN-Arxiv and 128 for others. 
For Sim-PGCN, we tune the weighting parameter $\lambda$ is searched from $\{0.1, 0.5, 1, 5, 10, 50, 100\}$ and $\gamma$ is searched from $\{0.01, 0.1\}$. 
For STABLE, we tune the Jaccard similarity threshold $t_1$ from $\{0.0, 0.01, 0.02, 0.03, 0.04, 0.05\}$, $k$ is tuned from $\{1, 3, 5, 7, 11, 13\}$, $\alpha$ is tuned from $-0.5$ to $3$. 
For HANG\_quad, we tune the time from $\{3, 6, 8, 15, 20, 25\}$, the hidden size from $\{16, 32, 64, 128\}$, and dropout from $\{0.2, 0.4, 0.6, 0.8\}$. 
For all experiments, we select the optimal hyper-parameters on the validation set and apply them to the test set. 

\subsection{Computing Environment and Resource}\label{exp: resource_cost}
The implementation of LLM4RGNN utilized the PyG module. With GPT-4, the token costs for constructing different sizes of instruction datasets are reported in Table~\ref{tab: token_cost}, and it is "spend once, use forever". The minimum resource requirements for LLMs are as follows: The 7B, 8B and 13B LLMs require 16G, 20G, and 32G for fine-tuning, and 15G, 16G, and 26G for deploying. The experiments are conducted in a computing environment with the following specifications: 
\begin{itemize}[leftmargin=*]
    \item \textbf{OS}: Linux ubuntu 5.15.0-102-generic. 
    \item \textbf{CPU}: Intel(R) Xeon(R) Platinum 8358 CPU @ 2.60GHz. 
    \item \textbf{GPU}: NVIDIA A800 80GB.
\end{itemize}
\begin{table}[tp]
\centering
\caption{The token cost of different instruction datasets.}\label{tab: token_cost}
\vskip -0.15in
\begin{tabular}{c|c|c|c}
\hline
\textbf{Size} & \textbf{Input Token} & \textbf{Output Token} & \textbf{Cost} \\ \hline
$5,000$  & $2,988,658$   & $561,762$      & $\$46.7$  \\ \hline
$15,000$ & $8,951,923$   & $1,681,254$    & $\$139.9$ \\ \hline
$26,518$ & $15,809,910$  & $2,970,500$    & $\$247.2$ \\ \hline
\end{tabular}
\vskip -0.2in
\end{table}
\begin{figure*}[htbp]
    \centering
    \subfigure[\textbf{Clean Cora.}]
{
    \begin{minipage}[b]{.22\linewidth}
        \centering
        \includegraphics[width=\textwidth, trim={0 0 0 2.5cm}, clip]{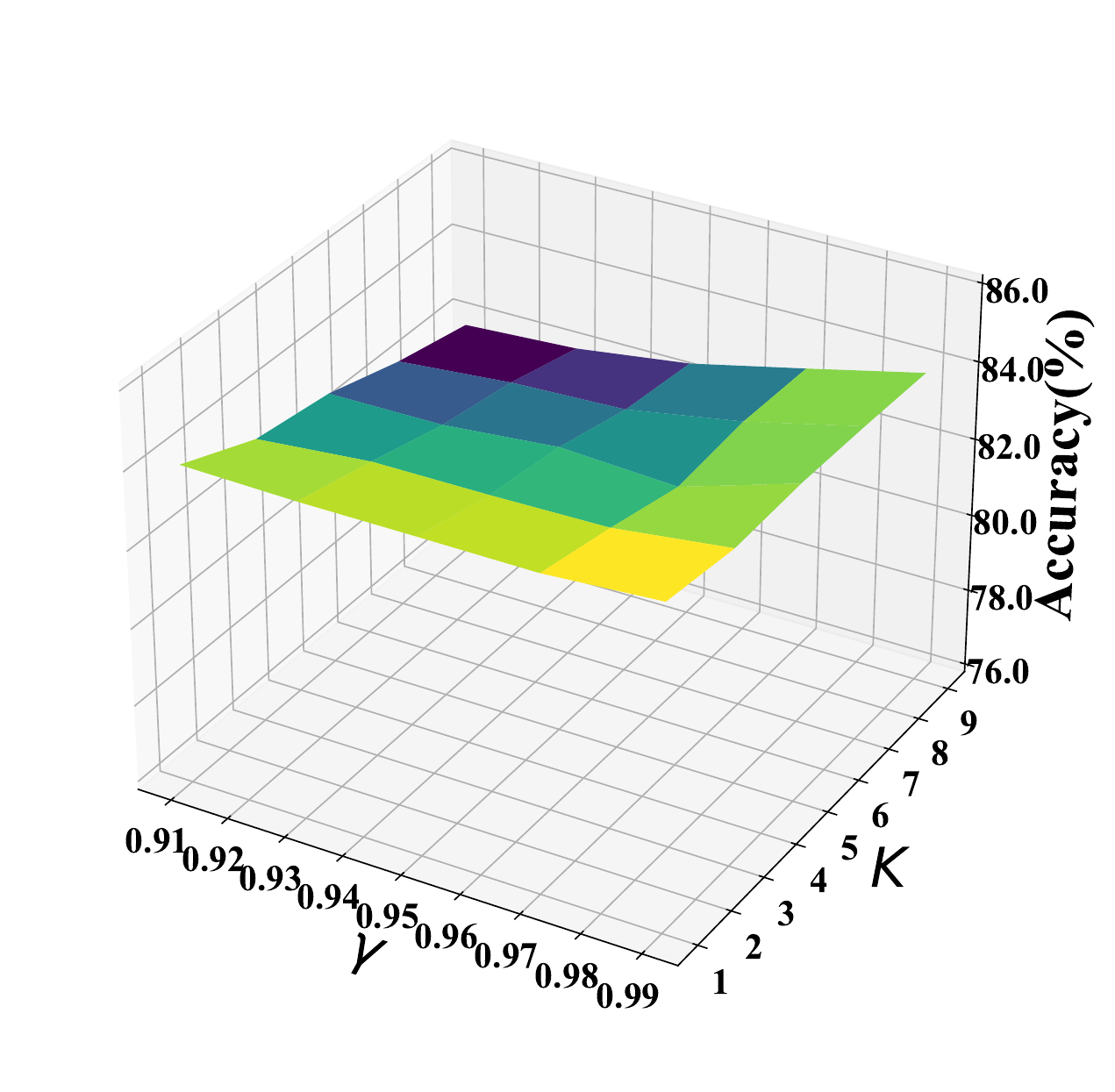}
        \vspace{-22pt}
    \end{minipage}
}
\subfigure[\textbf{Cora-Mettack-10\%.}]
{
 	\begin{minipage}[b]{.22\linewidth}
        \centering
        \includegraphics[width=\textwidth, trim={0 0 0 2.5cm}, clip]{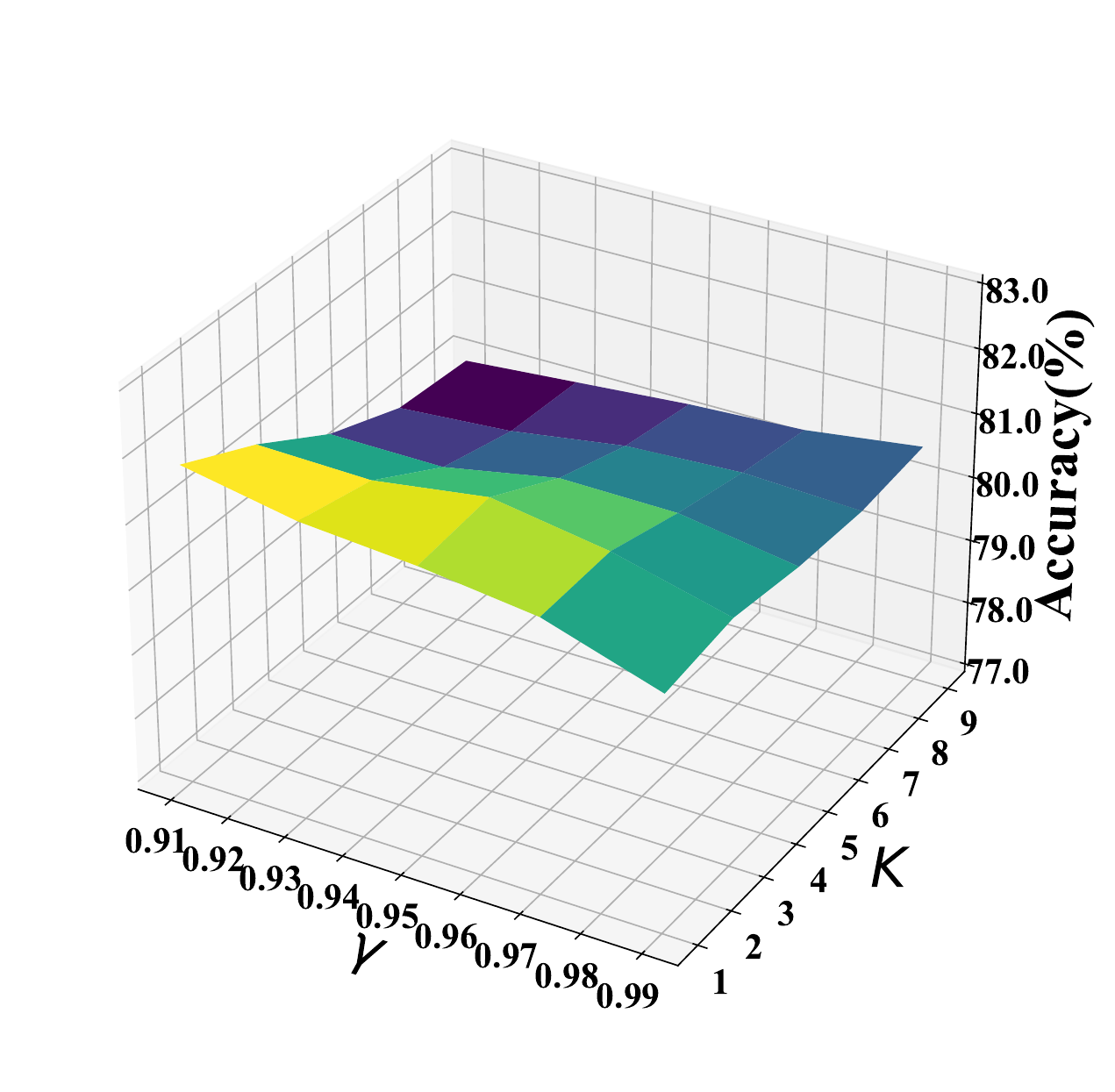}
        \vspace{-22pt}
    \end{minipage}
}
    \subfigure[\textbf{Clean Citeseer.}]
{
    \begin{minipage}[b]{.22\linewidth}
        \centering
        \includegraphics[width=\textwidth, trim={0 0 0 2.5cm}, clip]{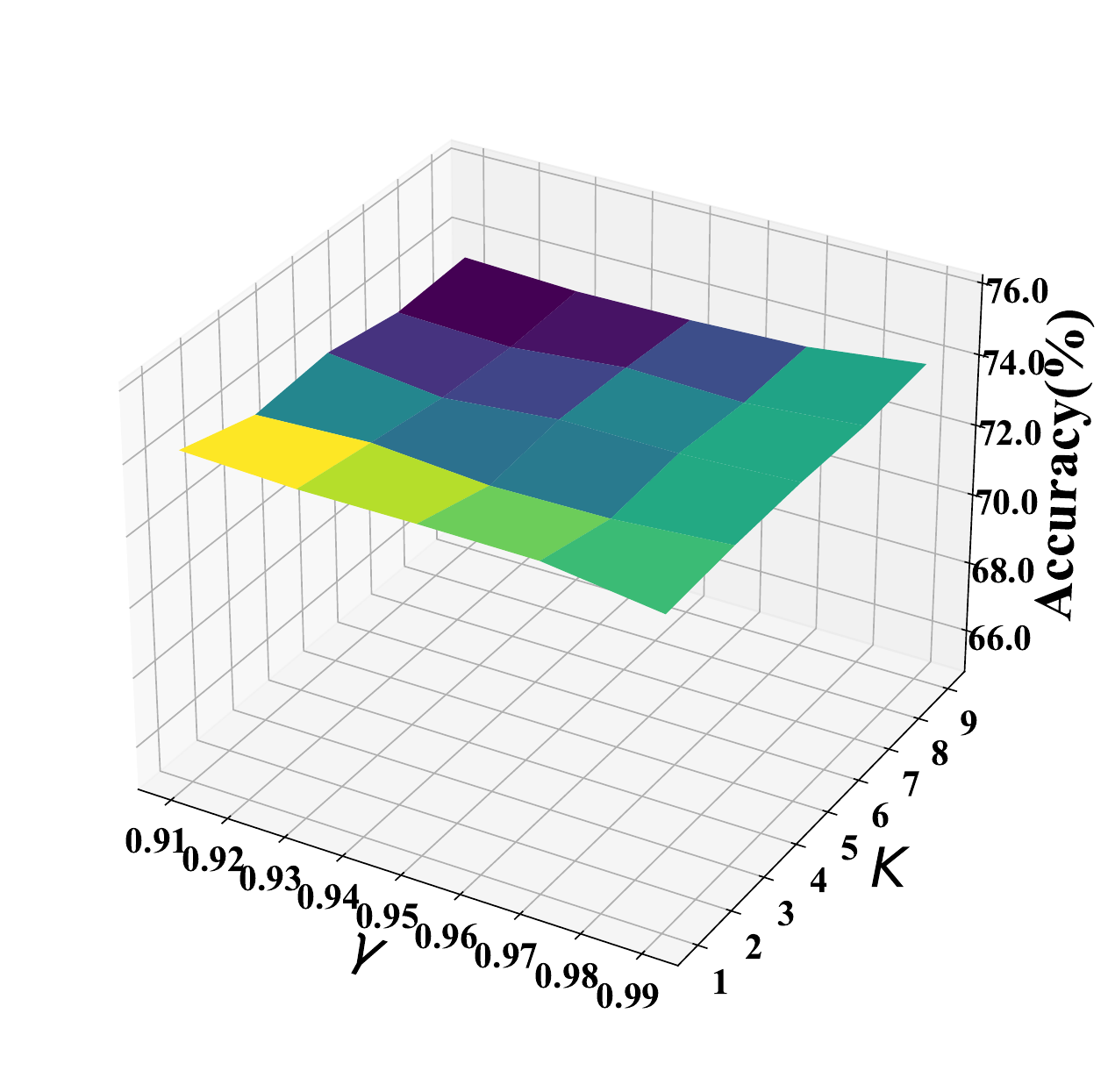}
        \vspace{-22pt}
    \end{minipage}
}
\subfigure[\textbf{Citeseer-Mettack-10\%.}]
{
 	\begin{minipage}[b]{.22\linewidth}
        \centering
        \includegraphics[width=\textwidth, trim={0 0 0 2.5cm}, clip]{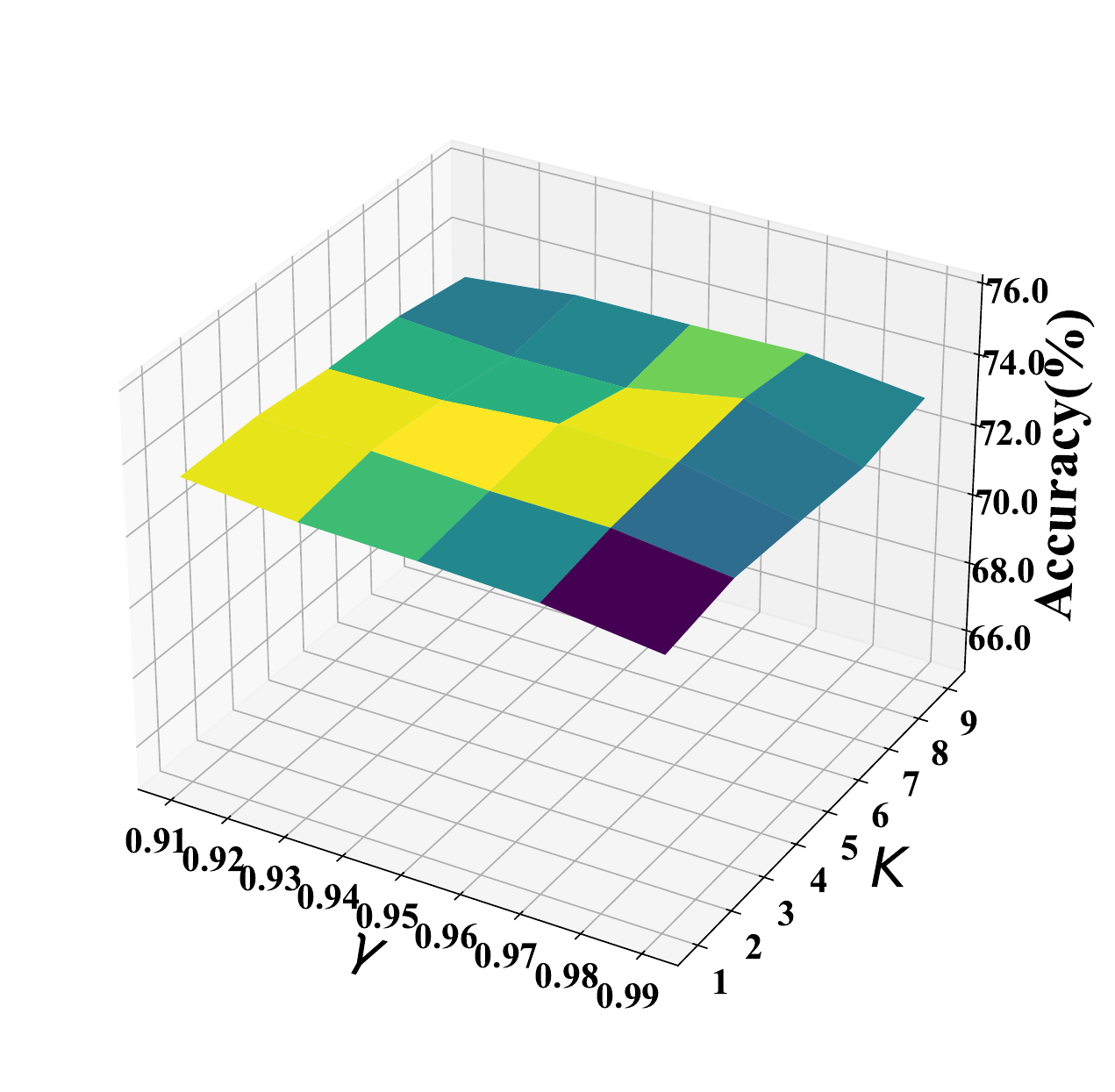}
        \vspace{-22pt}
    \end{minipage}
}
    \vskip -0.15in
    \caption{Analysis of the hyper-parameter $\gamma$ and $K$ against Mettack.}\label{fig: hyper_gamma_k_cora_citeseer_0_10}
    \vskip -0.15in
\end{figure*}
\balance
\section{More Experiment Result}
\subsection{Defense against Inductive Poisoning Attack}\label{exp: ind_poison}
We further evaluate the generalization of LLM4RGNN under inductive poisoning attacks. We conduct inductive experiments with Mettack on the Cora and Citeseer. Specifically, we randomly split the data into training, validation, and test sets with a 1:8:1 ratio. During training, we ensure the removal of test nodes and their connected edges from the graph. We perform Mettack on the validation set, purify the attacked graph using LLM4RGNN, and use the purified graph to train GNNs. The trained GNNs are then predicted on the clean test set. As reported in Table~\ref{tab: ind_poison_meta}, where we only report the baselines that support the inductive setting, results show that under the inductive setting, LLM4RGNN not only consistently improves the robustness of various GNNs but also surpasses the robust GCN framework HANG-quad, demonstrating its superior defensive capability in the inductive setting.
\begin{table}[tp]
\caption{Node classification accuracy ($\%\pm\sigma$) under Inductive Mettack. The best results are in bold.}
\vskip -0.15in
\Huge
\label{tab: ind_poison_meta}
\resizebox{\linewidth}{!}{
\centering
\begin{tabular}{p{0pt}c|c|cc|cccc} 
\hline
& \multicolumn{1}{c|}{\multirow{2}{*}{\makecell{Dataset \\ Ptb Rate}}} & \multicolumn{1}{c|}{\multirow{2}{*}{\makecell{HANG-quad}}} & \multicolumn{2}{c|}{GCN} & \multicolumn{2}{c}{GAT} \\ 
\cline{4-7}
& & & \multicolumn{1}{c}{Vanilla} & \multicolumn{1}{c|}{LLM4RGNN} & \multicolumn{1}{c}{Vanilla} & \multicolumn{1}{c}{LLM4RGNN} \\ 
\hline
\multirow{4}{*}{\hspace{-15pt} \rotcell{\makebox[10pt][c]{Cora}}}     
& 0\%  & $82.09\pm0.40$ & $84.36\pm0.28$ & \cellcolor{gray!25}$84.25\pm0.16$ & $84.35\pm0.62$ & \cellcolor{gray!25}$\mathbf{84.39\pm0.61}$ \\
& 5\%  & $75.85\pm0.67$ & $77.35\pm1.56$ & \cellcolor{gray!25}$83.17\pm0.17$ & $81.20\pm0.93$ & \cellcolor{gray!25}$\mathbf{83.68\pm0.26}$ \\
& 10\% & $70.95\pm1.26$ & $72.79\pm1.49$ & \cellcolor{gray!25}$83.02\pm0.34$ & $78.55\pm1.12$ & \cellcolor{gray!25}$\mathbf{83.48\pm0.49}$ \\
& 20\% & $58.81\pm1.99$ & $61.56\pm2.64$ & \cellcolor{gray!25}$82.69\pm0.65$ & $67.65\pm2.69$ & \cellcolor{gray!25}$\mathbf{83.14\pm0.69}$ \\ 
\hline
\multirow{4}{*}{\hspace{-15pt} \rotcell{\makebox[10pt][c]{Citeseer}}} 
& 0\%  & $73.15\pm0.94$ & $73.76\pm0.34$ & \cellcolor{gray!25}$74.07\pm0.51$ & $73.84\pm0.57$ & \cellcolor{gray!25}$\mathbf{74.26\pm0.69}$ \\
& 5\%  & $71.50\pm0.55$ & $70.60\pm0.56$ & \cellcolor{gray!25}$\mathbf{73.35\pm0.35}$ & $72.47\pm0.82$ & \cellcolor{gray!25}$73.25\pm0.91$ \\
& 10\% & $69.42\pm0.64$ & $67.64\pm0.73$ & \cellcolor{gray!25}$72.68\pm0.41$ & $71.11\pm1.02$ & \cellcolor{gray!25}$\mathbf{73.25\pm0.40}$ \\
& 20\% & $67.90\pm1.21$ & $61.35\pm1.35$ & \cellcolor{gray!25}$72.54\pm0.36$ & $65.76\pm2.85$ & \cellcolor{gray!25}$\mathbf{73.32\pm0.52}$ \\ 
\hline
\end{tabular}}
\vskip -0.15in
\end{table}

\begin{table}[hp]
\vskip -0.1in
\caption{The performance of LLM4RGNN with different prompt variants. The best results are in bold.} \label{tab: prompt_variant}
\vskip -0.15in
\resizebox{\linewidth}{!}{
\begin{tabular}{c|cc|cc}
\hline
\multirow{2}{*}{Method} & \multicolumn{2}{c|}{Cora} & \multicolumn{2}{c}{Citeseer}                 \\ \cline{2-5} 
                        & ACC (↑)     & AdvEdge (↓) & ACC (↑)    & \multicolumn{1}{c}{AdvEdge (↓)} \\ \hline
Filter                  & $\mathbf{81.02_{\pm1.00}}$  & $\mathbf{114(2.16\%)}$  & $\mathbf{73.06_{\pm0.69}}$ & $\mathbf{108(2.56\%)}$                       \\
Auxiliary Label                   & $80.80_{\pm1.07}$  & $165(3.13\%)$  & $72.17_{\pm0.75}$ & $121(2.86\%)$                       \\ \hline
\end{tabular}}
\vskip -0.2in
\end{table}
\subsection{The Impact of Different Prompt Variant}\label{exp: prompt_variant}
Recall that we utilize the ground truth of the query edge set to filter the incorrect prediction of GPT-4, there is another alternative is to directly take the ground truth as input of GPT-4. Specifically, we require GPT-4 to provide analysis and edge ratings ranging from 4 to 6 for a non-malicious edge, while providing edge ratings ranging from 1 to 3 for a malicious edge. Considering an instruction dataset with a size of 7500, we compared the results of the aforementioned method (Auxiliary Label) with our method (Filter). As reported in Table~\ref{tab: prompt_variant}, the performance variations of LLM4RGNN are minimal, demonstrating that LLM4RGNN is not sensitive to prompt design. Moreover, the reason why the lower performance of Auxiliary Label is that the edge label may be considered by GPT-4 during the generating analysis, but it is inaccessible during testing.

\begin{table}[hp]
\caption{Performance comparison between Mistral-7B and Llama3-8B under Mettack. The best results are in bold.}
\vskip -0.15in
\label{tab: llama-mistral}
\centering
\Huge
\setlength{\extrarowheight}{1.25pt}
\resizebox{\linewidth}{!}{
\begin{tabular}{p{0pt}c|cc|cc|cc} 
\hline
\multirow{2}{*}{} &
\multicolumn{1}{c|}{\multirow{2}{*}{\makecell{Dataset \\ Ptb Rate}}} & 
\multicolumn{2}{c|}{AdvEdge (↓)} & 
\multicolumn{2}{c|}{ACC (↑) w/o EP} & 
\multicolumn{2}{c}{ACC (↑) Full}  \\ 
\cline{3-8}
& & Mistral-7B & Llama3-8B & Mistral-7B & Llama3-8B & Mistral-7B & Llama3-8B \\ 
\hline

\multirow{4}{*}{\hspace{-15pt}\rotcell{\makebox[10pt][c]{Cora}}}                        & $0\%$                       & -           & -              & $\mathbf{83.78_{\pm 0.38}}$ & $83.72_{\pm 0.30}$                    & $84.13_{\pm 0.33}$ & $\mathbf{84.34_{\pm 0.56}}$                   \\
                                             & $5\%$                       & $\mathbf{28(0.53\%)}$  & $30(0.57\%)$     & $80.48_{\pm 0.19}$ & $\mathbf{80.60_{\pm 0.27}}$                    & $81.76_{\pm 0.69}$ & $\mathbf{81.98_{\pm 0.55}}$                   \\
                                             & $10\%$                      & $\mathbf{60(1.14\%)}$  & $63(1.19\%)$     & $79.83_{\pm 0.43}$ & $\mathbf{79.96_{\pm 0.39}}$                    & $\mathbf{81.80_{\pm 0.76}}$ & $81.77_{\pm 0.44}$                   \\
                                             & $20\%$                      & $\mathbf{102(1.93\%)}$ & $107(2.03\%)$    & $78.68_{\pm 0.28}$ & $\mathbf{79.22_{\pm 0.34}}$                    & $81.41_{\pm 0.77}$ & $\mathbf{81.71_{\pm 0.51}}$                   \\
\hline
\multirow{4}{*}{\hspace{-15pt}\rotcell{\makebox[10pt][c]{Citeseer}}}                    & $0\%$                       & -           & -              & $73.39_{\pm 0.69}$ & $\mathbf{73.41_{\pm 0.65}}$                    & $\mathbf{74.20_{\pm 0.56}}$ & $73.84_{\pm 0.50}$                   \\
                                             & $5\%$                       & $30(0.71\%)$  & $\mathbf{25(0.59\%)}$     & $72.17_{\pm 0.57}$ & $\mathbf{72.44_{\pm 0.52}}$                    & $\mathbf{73.94_{\pm 0.56}}$ & $73.28_{\pm 1.19}$                   \\
                                             & $10\%$                      & $55(1.30\%)$  & $\mathbf{47(1.11\%)}$     & $\mathbf{71.50_{\pm 0.59}}$ & $71.32_{\pm 0.58}$                    & $\mathbf{73.62_{\pm 0.39}}$ & $73.20_{\pm 0.68}$                   \\
                                             & $20\%$                      & $92(2.18\%)$  & $\mathbf{91(2.15\%)}$     & $\mathbf{71.07_{\pm 0.35}}$ & $70.85_{\pm 0.64}$                    & $\mathbf{74.12_{\pm 0.85}}$ & $74.06_{\pm 0.78}$                   \\
\hline
\end{tabular}}
\vskip -0.15in
\end{table}

\subsection{More Hyper-parameter Sensitivity}\label{sec: exp-hyper-details}
We conduct more hyper-parameter experiments on the probability threshold $\lambda$ and the number of important edges $K$. As shown in Figure~\ref{fig: hyper_gamma_k_cora_citeseer_0_10}, results indicate that the performance of the LLM4RGNN is stable under various parameter configurations and consistently outperforms existing SOTA methods.

\subsection{Comparison of Llama3-8B vs. Mistral-7B}\label{app: llama_mistral_compare}
In Section~\ref{exp: diff_llm}, Llama3-8B and Mistral-7B achieved close performance. Given that Llama3-8B is released after all experiments are completed, we conduct a comprehensive comparison between them here. As reported in Table~\ref{tab: llama-mistral}, both well-tuned LLMs can effectively identify malicious edges, and the performance gap is negligible.
\section{Case Study}\label{app: case_study}
LLMs have powerful understanding and inference abilities to complex text, thereby can effectively infer edge relations with texts. Here, we show some cases using GPT-4 and well-tuned Mistral-7B to infer the relationships between nodes. We find that the well-tuned Mistral-7B possesses the edge relation inference ability of GPT-4. They infer the edge relations and provide analysis by discussing the background, problems, methods, and applications of the two nodes.

\subsection{TAPE-Arxiv23 (GPT-4)}\label{case: arxiv}
\begin{tcolorbox}
\textbf{User content:} 
Node~1$\rightarrow$Title: when renewable energy meets building thermal mass a real time load management scheme\mynl Abstract: We consider the optimal power management in renewable driven smart building MicroGrid under noise corrupted conditions as a stochastic optimization problem. We first propose our user satisfaction and electricity consumption balanced (USECB) profit model as the objective for optimal power management. We then cast the problem in noise corrupted conditions into the class of expectation maximizing in stochastic optimization problem with convex constraints. For this task, we design a Bregemen projection based mirror decent algorithm as an approximation solution to our stochastic optimization problem. Convergence and upper-bound of our algorithm with proof are also provided in our paper. We then conduct a broad type of experiment in our simulation to test the justification of our model as well as the effectiveness of our algorithm.\\
Node~2$\rightarrow$Title: hiding sound in image by k lsb mutation\mynl Abstract: In this paper a novel approach to hide sound files in a digital image is proposed and implemented such that it becomes difficult to conclude about the existence of the hidden data inside the image. In this approach, we utilize the rightmost k-LSB of pixels in an image to embed MP3 sound bits into a pixel. The pixels are so chosen that the distortion in image would be minimized due to embedding. This requires comparing all the possible permutations of pixel values, which may would lead to exponential time computation. To speed up this, Cuckoo Search (CS) could be used to find the most optimal solution. The advantage of using proposed CS is that it is easy to implement and is very effective at converging in relatively less iterations/generations.
\tcblower
\textbf{Analysis:} The two nodes represent research in different domains: Node1 discusses renewable energy and building thermal mass in the context of smart grids, while Node2 focuses on hiding sound files in images using k-LSB mutation and Cuckoo Search. There is no direct connection between the optimization of renewable energy in buildings and the embedding of sound files in images. \textcolor{red}{The topics are unrelated, and the research methods and applications are distinct.}
\\
\textbf{Relevance Score:} 1
\end{tcolorbox}
\begin{tcolorbox}
\textbf{User content:} 
Node~1$\rightarrow$Title: a multilingual view of unsupervised machine translation\mynl Abstract: We present a probabilistic framework for multilingual neural machine translation that encompasses supervised and unsupervised setups, focusing on unsupervised translation. In addition to studying the vanilla case where there is only monolingual data available, we propose a novel setup where one language in the (source, target) pair is not associated with any parallel data, but there may exist auxiliary parallel data that contains the other. This auxiliary data can naturally be utilized in our probabilistic framework via a novel cross-translation loss term. Empirically, we show that our approach results in higher BLEU scores over state-of-the-art unsupervised models on the WMT'14 English-French, WMT'16 English-German, and WMT'16 English-Romanian datasets in most directions. In particular, we obtain a +1.65 BLEU advantage over the best-performing unsupervised model in the Romanian-English direction.
\\
Node~2$\rightarrow$Title: explicit cross lingual pre training for unsupervised machine translation\mynl Abstract: Pre-training has proven to be effective in unsupervised machine translation due to its ability to model deep context information in cross-lingual scenarios. However, the cross-lingual information obtained from shared BPE spaces is inexplicit and limited. In this paper, we propose a novel cross-lingual pre-training method for unsupervised machine translation by incorporating explicit cross-lingual training signals. Specifically, we first calculate cross-lingual n-gram embeddings and infer an n-gram translation table from them. With those n-gram translation pairs, we propose a new pre-training model called Cross-lingual Masked Language Model (CMLM), which randomly chooses source n-grams in the input text stream and predicts their translation candidates at each time step. Experiments show that our method can incorporate beneficial cross-lingual information into pre-trained models. Taking pre-trained CMLM models as the encoder and decoder, we significantly improve the performance of unsupervised machine translation.
\tcblower
\textbf{Analysis:} Both nodes discuss advancements in unsupervised machine translation, focusing on the use of cross-lingual data and methods to improve translation accuracy. Node1 introduces a probabilistic framework for multilingual neural machine translation, including a novel cross-translation loss term. Node2 proposes a novel cross-lingual pre-training method for unsupervised machine translation, emphasizing the importance of explicit cross-lingual training signals. \textcolor{red}{The common themes of unsupervised machine translation, cross-lingual data, and the pursuit of higher translation accuracy suggest a high relevance between the two nodes.}\\
\textbf{Relevance Score:} 6
\end{tcolorbox}

\subsection{Cora (Mistral-7B)}\label{case: cora}
\begin{tcolorbox}
\textbf{User content:} 
Node~1$\rightarrow$Title: A Neural Network Model of Memory Consolidation\mynl Abstract: Some forms of memory rely temporarily on a system of brain structures located in the medial temporal lobe that includes the hippocampus. The recall of recent events is one task that relies crucially on the proper functioning of this system. As the event becomes less recent, the medial temporal lobe becomes less critical to the recall of the event, and the recollection appears to rely more upon the neocortex. It has been proposed that a process called consolidation is responsible for transfer of memory from the medial temporal lobe to the neocortex. We examine a network model proposed by P. Alvarez and L. Squire designed to incorporate some of the known features of consolidation, and propose several possible experiments intended to help evaluate the performance of this model under more realistic conditions. Finally, we implement an extended version of the model that can accommodate varying assumptions about the number of areas and connections within the brain and memory capacity, and examine the performance of our model on Alvarez and Squire's original task.\\
Node~2$\rightarrow$Title: A Preliminary Investigation of Evolution as a Form Design Strategy\mynl Abstract: We describe the preliminary version of our investigative software, GGE Generative Genetic Explorer, in which genetic operations interact with Au-toCAD to generate novel 3D forms for the architect. GGE allows us to asess how evolutionary algorithms should be tailored to suit Architecture CAD tasks.
\tcblower
\textbf{Analysis:} The two nodes represent research in different domains: Node1 discusses a neural network model of memory consolidation in the brain, while Node2 focuses on a software tool for generating 3D forms in architecture using evolutionary algorithms. There is no direct connection between the brain's memory consolidation process and the evolution of 3D forms in architecture. \textcolor{red}{The topics are unrelated, and the expertise required for each is distinct.}
\\
\textbf{Relevance Score:} 1
\end{tcolorbox}

\begin{tcolorbox}
\textbf{User content:} 
Node~1$\rightarrow$Title: Constructive Algorithms for Hierarchical Mixtures of Experts\mynl Abstract: We present two additions to the hierarchical mixture of experts (HME) architecture. We view the HME as a tree structured classifier. Firstly, by applying a likelihood splitting criteria to each expert in the HME we "grow" the tree adaptively during training. Secondly, by considering only the most probable path through the tree we may "prune" branches away, either temporarily, or permanently if they become redundant. We demonstrate results for the growing and pruning algorithms which show significant speed ups and more efficient use of parameters over the conventional algorithms in discriminating between two interlocking spirals and classifying 8-bit parity patterns.
\\
Node~2$\rightarrow$Title: CLASSIFICATION USING HIERARCHICAL MIXTURES OF EXPERTS\mynl Abstract: There has recently been widespread interest in the use of multiple models for classification and regression in the statistics and neural networks communities. The Hierarchical Mixture of Experts (HME) has been successful in a number of regression problems, yielding significantly faster training through the use of the Expectation Maximisation algorithm. In this paper we extend the HME to classification and results are reported for three common classification benchmark tests: Exclusive-Or, N-input Parity and Two Spirals.
\tcblower
\textbf{Analysis:} Both nodes discuss the Hierarchical Mixture of Experts (HME) architecture, with Node1 focusing on constructive algorithms for HME and Node2 on extending HME to classification tasks. The abstracts of both nodes mention the use of HME in classification problems, indicating a direct relevance in the context of machine learning and classification techniques. \textcolor{red}{The shared focus on HME and its application in classification problems suggests a high relevance between the two nodes.}
\\
\textbf{Relevance Score:} 6
\end{tcolorbox}

\end{document}